\documentclass[10pt,twocolumn,letterpaper]{article}

\usepackage{3dv}
\usepackage{times}
\usepackage{epsfig}
\usepackage{graphicx}
\usepackage{amsmath}
\usepackage{amssymb}

\usepackage{xcolor}
\usepackage{enumitem}
\usepackage{booktabs}
\usepackage{multicol}
\usepackage{multirow}
\usepackage{subcaption}
\usepackage{graphicx}
\usepackage{mathtools}
\usepackage{siunitx}
\usepackage{booktabs}
\usepackage{etoolbox}

\newcommand{\R}{\mathbb{R}}

\renewcommand{\epsilon}{\ensuremath\varepsilon}

\renewcommand{\phi}{\ensuremath{\varphi}}

\newcommand{\pc}[1]{\ensuremath{\mathcal{X}_{#1}}} % Point cloud 1

\newcommand{\boldparagraph}[1]{\vspace{0.1em}\noindent{\bf #1} }
\newcommand{\sfvec}[1]{{d}_{#1}}
\newcommand{\pcpoint}[1]{{x}_{#1}}
\newcommand{\ofvec}[1]{{v}_{#1}}
\newcommand{\campose}[1]{T_{#1}}

\DeclarePairedDelimiter\norm{\lVert}{\rVert}

\newcommand{\hatsub}[1]{\expandafter\hat#1}

\newcommand{\ra}[1]{\renewcommand{\arraystretch}{#1}}
\newcommand{\lossterm}[1]{\ensuremath{\mathcal{L}_{#1}}}

\usepackage[pagebackref=true,breaklinks=true,letterpaper=true,colorlinks,bookmarks=false]{hyperref}

\threedvfinalcopy % *** Uncomment this line for the final submission

\def\threedvPaperID{219} % *** Enter the 3DV Paper ID here

\begin{document}

%%%%%%%%% TITLE
\title{Self-Supervised Learning of Non-Rigid Residual Flow and Ego-Motion}

\author{Ivan Tishchenko$^1$
% ETH Zurich\\
\qquad
Sandro Lombardi$^1$
\qquad
Martin R. Oswald$^1$
\qquad
Marc Pollefeys$^{1,2}$
\\ $^1$Department of Computer Science, ETH Zurich \qquad $^2$Mixed Reality and AI Zurich Lab, Microsoft
\\ {\tt\small firstname.lastname@inf.ethz.ch}
}

\maketitle
%%%%%%%%% ABSTRACT
\begin{abstract}
Most of the current scene flow methods choose to model scene flow as a per point translation vector without differentiating between static and dynamic components of 3D motion. In this work we present an alternative method for end-to-end scene flow learning by joint estimation of non-rigid residual flow and ego-motion flow for dynamic 3D scenes. 
We propose to learn the relative rigid transformation from a pair of point clouds followed by an iterative refinement. We then learn the non-rigid flow from transformed inputs with the deducted rigid part of the flow. Furthermore, we extend the supervised framework with self-supervisory signals based on the temporal consistency property of a point cloud sequence. Our solution allows both training in a supervised mode complemented by self-supervisory loss terms as well as training in a fully self-supervised mode. We demonstrate that decomposition of scene flow into non-rigid flow and ego-motion flow along with an introduction of the self-supervisory signals allowed us to outperform the current state-of-the-art supervised methods.
\end{abstract}

%%%%%%%%% BODY TEXT
\section{Introduction}
In order to build fully autonomous agents we need to enable them to perceive the world in 3D and also to do reasoning about it directly in 3D.
Scene flow estimation is crucial for understanding 3D motion and is an essential building block to advance emerging technologies in areas such as dynamic reconstruction, robotic perception, autonomous driving \etc.

%Deep learning based methods \cite{gu2019hplflownet, lv2018learning} have been successfully used for supervised scene flow estimation.
Deep learning based scene flow methods \cite{gu2019hplflownet, lv2018learning} often rely on supervised training. 
Hence, one common property of most learning based methods is that they require large amounts of labelled observations for training. Acquiring the ground truth of scene flow for real-world data is laborious or sometimes impossible. As a consequence, researches have been relying on using rendered synthetic datasets \cite{mayer2016large, lv2018learning, Menze2015ISA, Menze2015CVPR} for training. This spawns a discrepancy between real-world and synthetic data. Concretely, networks trained on synthetic data are not guaranteed to generalize to raw real-world data. The reason for a gap is that synthetic datasets are not always shape and depth realistic and that they generally capture one specific domain.
One way to tackle these issues is to utilize large amounts of unlabeled data with self-supervised learning.

Self-supervised learning has great potential for 3D motion estimation because acquiring real-world ground truth for 3D data is especially laborious. Self-supervised learning has shown its effectiveness in bridging the gap between synthetic and real-world data for the case of optical flow~\cite{liu2019selflow}. Additionally, the combination of supervised learning and self-supervision is possible with supervised pre-training and fully self-supervised fine-tuning. Furthermore, one can design self-supervised losses which serve as implicit regularizers for the supervised part of the loss. This kind of configuration can also be useful in a weakly supervised setting, where the method can still learn from datasets with incomplete labels and put more weight on the self-supervised part of the loss if the labels in the dataset are inaccurate.

Learning based approaches for scene flow \cite{gu2019hplflownet, liu2019flownet3d} typically model scene flow holistically as a 3D translation vector without any distinct separation of camera and object motion. While this type of scene flow modeling is acceptable in case of static scenes, it is beneficial for the case of dynamic scenes to deduce which part of the scene flow is induced by the ego-motion of an observer and which part is induced by the object itself. In this case one single inference pass allows us to obtain pure object motion, relative camera pose and the total motion of a dynamic 3D scene.

Our work complements previous works on scene flow learning with 3D deep learning \cite{gu2019hplflownet, su2018splatnet, liu2019flownet3d} by introducing self-supervision and an alternative approach for learning scene flow which disentangles ego-motion from the object motion in contrast to approaches which rely on total scene flow only \cite{mittal2020just, wu2019pointpwc, wang2020flownet3d++}. While there have been works on separating the camera motion and object motion for scene flow, most of them learn on 2.5D datasets \cite{lv2018learning, behl2019pointflownet, liu2019unsupervised}, which makes them inapplicable if raw data is acquired in the point cloud format directly. Moreover, our approach aims at an end-to-end self-supervised learning pipeline in contrast to learning different parts of the flow with separate components.
In summary, this work's \textbf{contributions} are:
\begin{itemize}[itemsep=1pt,topsep=0pt,leftmargin=*]
\item An end-to-end joint learning of ego-motion and non-rigid flow with 3D deep learning from point clouds by extending a current state of art total flow architecture HPLFlowNet~\cite{gu2019hplflownet}.
\item An alternative approach for scene flow learning which decomposes the total flow into non-rigid and rigid parts. Comparison with a traditional motion model of holistic total scene flow.
\item Investigation of self-supervision for non-rigid flow and ego-motion learning using supervised training with self-supervisory signals and training in a fully self-supervised mode.
\end{itemize}

%------------------------------------------------------------------------
\section{Related Work}
\boldparagraph{Deep Scene Flow.}
Recent approaches \cite{liu2019flownet3d, gu2019hplflownet, wang2020flownet3d++, wu2019pointpwc} in scene flow estimation focused on estimating scene flow using supervised learning directly from point clouds with an end-to-end 3D deep learning architecture. To obtain large amounts of training data researchers have been employing synthetic stereo or RGB-D datasets \cite{mayer2016large, lv2018learning} to generate a point cloud representation of the scene.

A group of approaches \cite{wang2020flownet3d++, mittal2020just} build on top of FlowNet3D proposed by Liu~\etal~\cite{liu2019flownet3d}, which is based on point encoders developed by Qi~\etal~\cite{qi2017pointnet, qi2017pointnet++}. These type of networks encode each point or alternatively a neighbourhood of points individually using a shared multi-layer perceptron (MLP) and then form a global aggregated feature by using global pooling. The shared characteristic of these methods is that they loose some degree of structural information between points due to separate point encoding followed by aggregation with global max pooling.

Gu \etal~\cite{gu2019hplflownet} proposed a supervised learning approach HPLFlowNet which directly estimates scene flow from point clouds using a permutohedral lattice network. 
Permutohedral lattice networks \cite{su2018splatnet, gu2019hplflownet} are able to efficiently handle point clouds with large number of points due to their high dimensional sparse representation of the input signal. Furthermore, in contrast to learning on regular integer grids, permutohedral lattice networks promote scalability in input dimensions $d > 3$ if one is to include RGB, intensity or normals information in addition to spatial location vector. We refer the reader to a seminal work by Adams~\etal~\cite{adams2010fast} for further properties of permutohedral lattices.

\boldparagraph{Self-Supervised Learning.}
Self-supervision for scene flow learning is an active field of research and there have been concurrent works \cite{mittal2020just, wu2019pointpwc, wang2020flownet3d++} to ours. However, these approaches assume non-rigidity of motion and thus do not differentiate between ego and object motion.

In their work, Wang~\etal~\cite{wang2019learning} have demonstrated how cycle-consistency of time can serve as a powerful self-supervisory signal for various visual correspondence tasks. Liu~\etal~\cite{liu2019flownet3d} proposed to use cycle-consistency for scene flow estimation in their self-supervised loss term along with a supervised loss. Our method is inspired by findings of Mittal~\etal~\cite{mittal2020just}, who proposed to complement the idea of cycle-consistency \cite{liu2019flownet3d, wang2019learning} with the nearest neighbour anchoring to eliminate degenerate cases of zero scene flow predictions. Furthermore, they pose a nearest neighbour constraint to bring scene flow-warped points of the first point cloud close to the point of the second point cloud to complement cycle consistency. 

Wang~\etal~\cite{wang2020flownet3d++} used the nearest neighbour constraint in form of point-to-plane error, while Wu \etal~\cite{wu2019pointpwc} extended the nearest neighbour constraint to both forward and backward directions with Chamfer distance loss.

\boldparagraph{Relative Camera Pose Estimation.}
Most of scene flow methods \cite{gu2019hplflownet, liu2019flownet3d, wu2019pointpwc, mittal2020just} choose to model scene flow as a 3D translation vector. An alternative view on scene flow disentangles camera motion from the object motion in dynamic 3D scenes. 
Behl~\etal~\cite{behl2019pointflownet} proposed to jointly learn a per-point rigid transformation along with rigid ego-motion and bounding boxes from point clouds using a single neural network.
Lv~\etal~\cite{lv2018learning} showed that modelling camera motion as a rigid transformation and the rest of the motion as non-rigid per point translations improves non-rigid flow estimation on RGB-D inputs.

Point cloud inputs for scene flow estimation contain some initial alignment. Therefore, relative camera transformation can be estimated by using local point cloud registration. We follow the approaches \cite{yuan2018iterative, aoki2019pointnetlk, gojcic2020learning, choy2020deep} that have shown that iterative point cloud registration with 3D deep neural networks can accurately estimate a rigid transformation between point clouds. Yuan~\etal~\cite{yuan2018iterative} proposed iterative transformer networks to refine the rigid transform. In their work they estimate an initial rigid transformation with a point based network and then apply that transformation to the inputs. Afterwards, the same process is repeated $K$ times. At the end they obtain a final pose, which is a matrix product of $K$ intermediate steps. Aoki~\etal~\cite{aoki2019pointnetlk} proposed a recurrent deep learning version of the LK method~\cite{lucas1981iterative}.
We provide an overview of related work in Table~\ref{table:related}.
% \boldparagraph{Method Comparison}.
% We present a method comparison with related work in Table \ref{table:related}.
\begin{table}[!htbp]
    \ra{1.2}
    \begin{center}
    \resizebox{\columnwidth}{!}{
        \begin{tabular}{@{}l l c c c@{}}
            \toprule
            Approach & Format & Self. & Ego-motion & End-to-end\\
            \midrule
            Learning Rigidity \cite{lv2018learning} & RGB-D  &  & \checkmark &  \\
            PointFlowNet \cite{behl2019pointflownet} & Point clouds  &  & \checkmark & \checkmark \\
            % \cmidrule{1-5}
            HPLFlowNet \cite{gu2019hplflownet} & Point clouds &  &  & \checkmark \\
            FlowNet3D \cite{liu2019flownet3d} & Point clouds  &  &  & \checkmark \\
            FlowNet3D++ \cite{wang2020flownet3d++} & Point clouds &  &  & \checkmark \\
            % \cmidrule{1-5}
            PointPWC-Net \cite{wu2019pointpwc} & Point clouds  & \checkmark &  & \checkmark \\
            Just Go with the Flow \cite{mittal2020just} & Point clouds & \checkmark &  & \checkmark \\
            % \cmidrule{1-5}
            \textbf{Ours} & \textbf{Point clouds} & \textbf{\checkmark} & \textbf{\checkmark} & \textbf{\checkmark} \\
            \bottomrule
        \end{tabular}
    }
    \end{center}
    \vspace{-1.5em}
    \caption{\textbf{Scene flow methods overview.} We combine end-to-end learning of non-rigid flow and ego-motion from point clouds with self-supervision (denoted by "Self.").}
    \label{table:related}
\end{table}

%------------------------------------------------------------------------
\section{Method}
\boldparagraph{Problem Definition.}
Let $\pc{t} \in \R^{n \times d}$ be a point cloud representation of the scene at time step $t$ and $\pc{t+\delta} \in \R^{m \times d}$ be a point cloud at time step $t + \delta$. This pair of point clouds has the following properties:
\begin{itemize}[itemsep=1pt,topsep=0pt,leftmargin=*]
    \item Number of points $n$ and $m$ may be $n \neq m$.
    \item Points of point cloud sets $\pc{t}$ and $\pc{t+\delta}$ are not ordered.
    \item Point $\pcpoint{}$ in $\pc{t}$ may not be represented in $\pc{t+\delta}$.
\end{itemize}
Following \cite{gu2019hplflownet, su2018splatnet} each point $\pcpoint{}$ in the point cloud $\pc{}$ is associated with a position vector $\Vec{p} \in \R ^ {d}$ and a signal vector $\Vec{v} \in \R ^ {k}$. In the scope of this work our signal vectors equal to position vectors $\Vec{v} = \Vec{p}$ with dimensions $d = k = 3$.

Having $\pc{t}$ and $\pc{t+\delta}$ our main goal is to estimate total scene flow $\sfvec{} \in \R ^ 3$. Total scene flow is a per-point displacement vector which describes where a point from $\pc{t}$ will be in the next time step $t+\delta$.

\begin{figure}[!htbp]
    \def\svgwidth{\columnwidth}
    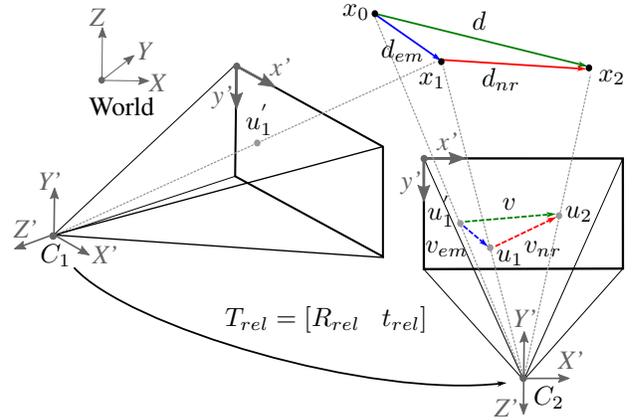
    \vspace{-0.8cm}
    \caption{\textbf{Full scene flow model.} Given the points $x_{0} \in \pc{t}$ and $x_{2} \in \pc{t+1}$, the total flow $\sfvec{}$ is a vector sum of ego-motion flow $\sfvec{em}$ and non-rigid flow $\sfvec{nr}$. $T_{rel}$ transforms a point of $\pc{t}$ from coordinate system $C_{1}$ into a coordinate system $C_{2}$ of $\pc{t+1}$. Optical flow $\ofvec{}$ is obtained by projecting the total flow vector $\sfvec{}$ into an image plane.}
    \label{figure:motion}
    \vspace{-0.1cm}
\end{figure}

\boldparagraph{3D Motion Model.}
As in \cite{quiroga2014dense, lv2018learning} our total flow $\sfvec{} \in \R^3$ consists of ego-motion flow $\sfvec{em} \in \R^3$ and non-rigid flow $\sfvec{nr} \in \R^3$: 
\begin{equation} \label{eq:total}
  \sfvec{} = \sfvec{nr} + \sfvec{em}
\end{equation}
The non-rigid flow $\sfvec{nr}$ represents the motion induced by a moving object. In contrast, ego-motion flow $\sfvec{em}$ describes per point motion induced by an observer.
The ego-motion is modeled with the rigid relative camera pose $T_{rel} = \begin{bmatrix}R_{rel} & t_{rel}\end{bmatrix} \in \R ^ {3 \times 4}$. $T_{rel}$ transforms a point $\pcpoint{t}$ from a coordinate system of $\pc{t}$ into a coordinate system of $\pc{t+\delta}$. 
Therefore, $\sfvec{em}$ is defined by:
\begin{equation} \label{eq:em}
    \sfvec{em} = \left(R_{rel} - I_{3\times 3}\right) \pcpoint{t} + t_{rel}
\end{equation}
Full scene flow model and its relation to optical flow is demonstrated in Figure \ref{figure:motion}.

\subsection{Network Architecture}
\label{sec:arch}
\begin{figure*}[!htbp]
    \begin{center}
        \resizebox{\linewidth}{!}{\includegraphics{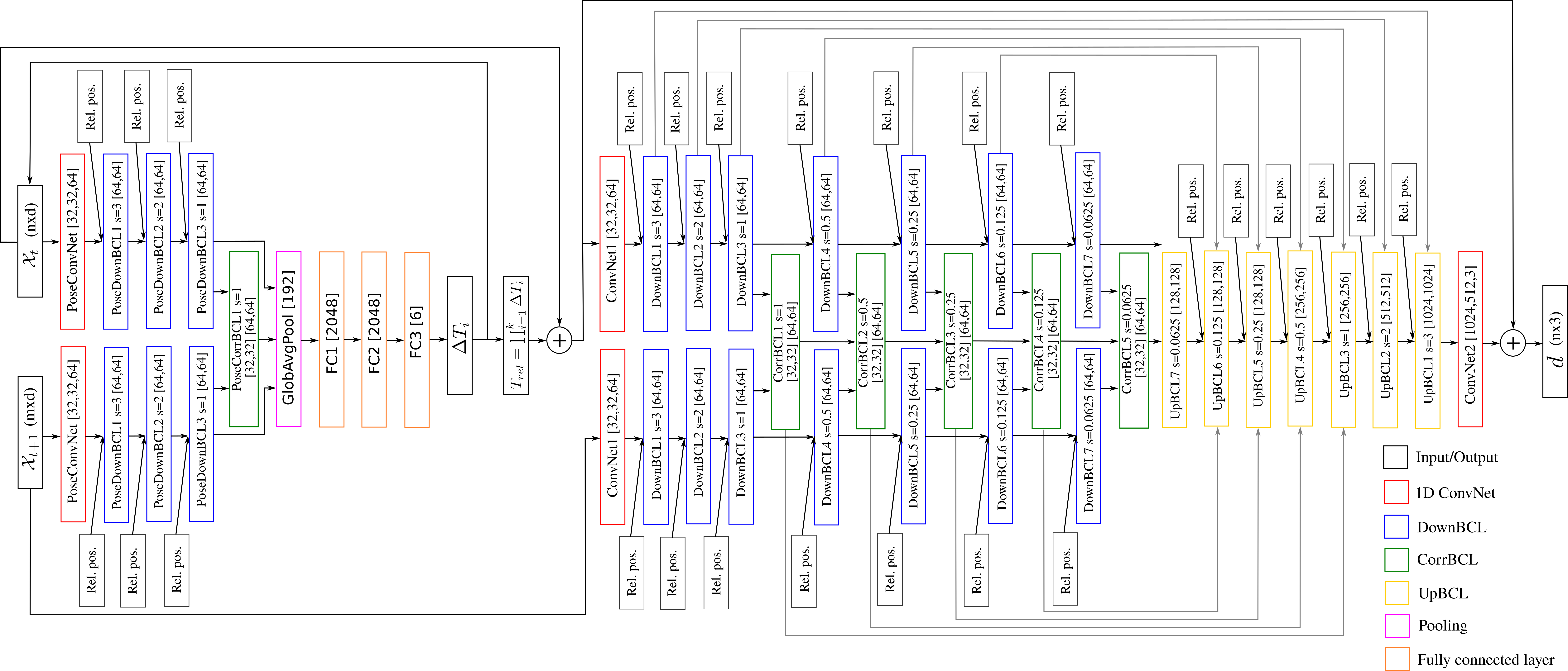}}
    \end{center}
    \vspace{-0.6cm}
    \caption{\textbf{Network architecture.} The network takes point clouds $\pc{1}$ and $\pc{2}$ as inputs. In the first stage a relative pose $T_{rel}$ is estimated in the pose regression network in order to obtain an ego-motion flow. To improve the estimation of a relative pose optional iterative refinement with $K$ steps is utilized. Pose transformed input $\pc{1}^{*}$ and $\pc{2}$ are fed into a method agnostic non-rigid flow network to learn the motion induced by a moving object. At the final stage, non-rigid flow and ego-motion flow are combined into a total scene flow.}
    \label{figure:arch}
\end{figure*}
This section introduces the pipeline that takes a pair of point clouds $\pc{1}$ and $\pc{2}$ and learns a relative camera pose.
To achieve a better pose estimate we optionally refine the pose in a recurrent fashion. 
Then, we transform the original input $\pc{1}$ with the estimate of the relative pose $\hat{T}_{rel}$ and feed it to the non-rigid flow network.
Finally, the estimated non-rigid flow and relative pose are combined into a total scene flow. Figure \ref{figure:arch} shows an overview of the proposed pipeline.

\boldparagraph{Relative Pose Regressor.}
We use DownBCL and CorrBCL \cite{gu2019hplflownet} as building blocks for our relative pose regression network which learns a relative pose $\hat{T}_{rel}$ directly from a pair of point clouds $\pc{1}$ and $\pc{2}$. 
The pose regression network consists of a pair of Siamese branches with shared weights. 
We limit the deepness of the pose regression network to 3 DownBCL blocks since we need coarse-level features for the relative pose. 
A pair of point clouds is down-sampled with DownBCLs and outputs of each branch are correlated inside a CorrBCL.
Afterwards, the branch outputs are concatenated together with a correlated feature to form one common representation. 

Since the number of output lattice points from DownBCL, CorrBCL is not fixed, we choose to perform global average pooling to aggregate a feature across all points in a channel. 
A 192-dimensional feature is fed into a sequence of two fully connected layers with 2048 neurons each.
We use Euler angles to represent the rotation since the optimization was more stable than with quaternions in practice.
Therefore, the output of a relative pose network is a 6-dimensional vector obtained from learning a translation $\hat{t} \in \R^{3}$ and Euler angles $\hat{r} \in \R^{3}$ together.

\boldparagraph{Iterative Pose Refinement.}
We introduce a refinement step to make the relative pose estimate more accurate and robust. 
Refinement is performed in a recurrent fashion inspired by deep iterative point set registration \cite{yuan2018iterative, aoki2019pointnetlk} which unfolds the estimation into $K \in [1, \infty)$ iterations.
The points of $\pc{1}$ are brought closer to the points of $\pc{2}$ with each iteration using an intermediate relative pose prediction.

The estimation starts with the initialization of the relative pose to an identity matrix $T_{rel} = \Delta T_{0} = I_{4\times 4}$.
Then, an intermediate relative pose $\Delta T_{i}$ is estimated at each of $K$ iterations with a network.
At each step $i$ the network's output $\Delta T_{i - 1}$ is used to transform the original $\pc{1}$ according to the recurrence $\pc{i}^{*} = \prod_{j=0}^{i - 1} \Delta T_{j} \pc{1}$ which is fed to the network and the same process is repeated in the next $i + 1 ... K$ steps.
After $K$ iterations we obtain a refined relative pose which is a matrix product of $K$ intermediate estimates $\hat{T}_{rel} = \prod_{i=1}^{k} \Delta T_{i}$.

The number of refinement iterations $K$ becomes a variable hyper-parameter.
Since the networks weights are shared between iterations, it is possible to trade off accuracy and runtime by setting a different $K$ for training and inference.
We refer the reader to \cite{yuan2018iterative, aoki2019pointnetlk} for a thorough investigation of how the hyper-parameter $K$ affects the performance in the context of point sets registration.

\boldparagraph{Non-Rigid Flow Network.}
After the relative pose estimation network's forward pass we obtain an estimate of the pose $\hat{T}_{rel} = \prod_{i=1}^{k} \Delta T_{i}$ and transformed original input of $\pc{1}$ defined by $\pc{1}^{*} = \prod_{i=1}^{k} \Delta T_{i} \pc{1}$. When there is no refinement .i.e. when $K = 1$, our transformed original input would just become $\pc{1}^{*} = \Delta T_{1} \pc{1}$. 

The idea in any case is to deduct the ego-motion from the inputs such that only the non-rigid flow can be learned in a non-rigid flow network. Our method is agnostic in terms of the non-rigid flow network. We chose to use HPLFlowNet \cite{gu2019hplflownet} for the task of learning non-rigid flow which is a state of the art supervised total flow learning approach.

Non-rigid flow network takes a transformed input $\pc{1}^{*}$ and $\pc{2}$ and learns the non-rigid flow $\hat{d}_{nr}$. An estimate of an ego-motion flow $\hat{d}_{em}$ is obtained with the relative pose $\hat{T}_{rel}$ estimated in the previous step according to the Equation (\ref{eq:em}). In the final step $\hat{d}_{nr}$ and $\hat{d}_{em}$ are summed to obtain a total flow estimate $\hat{d}$. 

\subsection{Loss functions}
\boldparagraph{Total Flow Loss.} The end point error (EPE3D) is a supervised loss which measures an average of the Euclidean distance over $n$ points between the ground truth total scene flow $d \in \R^{3}$ and the estimated total scene flow $\hat{d} \in \R^{3}$:
\begin{equation} \label{loss:epe3d}
    \lossterm{epe3d} = \frac{1}{n} \sum_{i=1}^{n} \norm{d_{i} - \hat{d}_{i}}_{2}
\end{equation}
where the estimate $\hat{d}$ is composed from the estimates of the rigid part $R, t$ and the non-rigid part $d_{nr}$ as:
\begin{equation}
    \hat{d} = \hat{d}_{nr} + (\hat{R} - I_{3\times 3})x_{1} + \hat{t}
\end{equation}

\boldparagraph{Non-Rigid Flow Loss.}
To estimate an error for non-rigid flow $\hat{d}_{nr} \in \R^{3}$ and the ground truth non-rigid flow $d_{nr} \in \R^{3}$ we use a similar loss as in Equation~\eqref{loss:epe3d}:
\begin{equation} \label{loss:nonrigid}
    \lossterm{nr} = \frac{1}{n} \sum_{i=1}^{n} \norm{(d_{nr})_i - (\hat{d}_{nr})_i}_{2}
\end{equation}

\boldparagraph{Rigid Loss.}
The rigid relative camera pose loss consist of a rotation and a translation term:
\begin{equation} \label{loss:rigid}
    \lossterm{r} = w_{rot} \lossterm{rot} + \lossterm{t}
\end{equation}
The rotation loss measures the distance between the predicted Euler angles $\hat{r} \in \R^{3}$ and the ground-truth angles $r \in \R^{3}$:
\begin{equation}
    \lossterm{rot} = \frac{1}{n} \sum_{i=1}^{n} \norm{r_{i} - \hat{r}_{i}}_{2}
\end{equation}
The translation loss $\lossterm{t}$ is the Euclidean distance between the ground truth relative translation $t \in \R^{3}$ and the prediction $\hat{t} \in \R^{3}$:
\begin{equation}
    \lossterm{t} = \frac{1}{n} \sum_{i=1}^{n} \norm{t_{i} - \hat{t}_{i}}_{2}
\end{equation}

\boldparagraph{Self-Supervised Loss.}
We employ self-supervised signals proposed by \cite{mittal2020just}.
The loss consists of a forward-backward consistency loss \lossterm{fb} and the nearest neighbour loss \lossterm{nn}:
\begin{equation} \label{loss:ss}
    \lossterm{ss} = w_{fb} \lossterm{fb} + w_{nn} \lossterm{nn}
\end{equation}
where \lossterm{fb} is the difference between all the points in the first time step $x_{1} \in \pc{1}$ and their corresponding alignments $\tilde{x}_{1} \in \R^{3}$ at the end of a forward-backward cycle:
\begin{equation}
    \lossterm{fb} = \frac{1}{|\pc{1}|} \sum_{x_{1} \in \pc{1}} \norm{x_{1} - \tilde{x}_{1}}_{2}
\end{equation}
An alignment $\tilde{x}_{1}$ is obtained by first warping the original point with a predicted forward flow $\tilde{x}_{2} = x_{1} + \hat{d}$.
Afterwards, a reverse flow is predicted from an anchor point $x_{2}^{*}$ into the original point cloud $\pc{1}$.
The anchor point $x_{2}^{*} = \frac{\tilde{x}_{2} + x_{nn}}{2}$ \cite{mittal2020just} is defined as the average between the warped point $\tilde{x}_{2}$ and its nearest neighbour $x_{nn} \in \pc{2}$.
Then, an anchor point $x_{2}^{*}$ is warped with a corresponding predicted reverse flow $\tilde{d}$ such that $\tilde{x}_{1} =  x_{2}^{*} + \tilde{d}$.

We adopt a nearest neighbour constraint \cite{mittal2020just} to stabilize the forward-backward consistency loss. 
The nearest neighbour loss aims at bringing all the forward flow warped points $\tilde{x}_{2} = x_{1} + \hat{d}$ close to their corresponding nearest neighbours $x_{nn} \in \pc{2}$. 
\begin{equation}
    \lossterm{nn} = \frac{1}{|\tilde{\pc{2}}|} \sum_{\tilde{x}_{2} \in \tilde{\pc{2}}} \norm{\tilde{x}_{2} - x_{nn}}_{2}
\end{equation}
\boldparagraph{Overall Loss.}
We combine all loss terms from Equations \eqref{loss:epe3d}, \eqref{loss:nonrigid}, \eqref{loss:rigid}, \eqref{loss:ss} into a weighted total loss.
\begin{equation} \label{loss:total2}
    \lossterm{} = w_{epe3d} \lossterm{epe3d} + w_{nr} \lossterm{nr} + w_{r} \lossterm{r} + \lossterm{ss}
\end{equation}
%
%------------------------------------------------------------------------
\section{Experiments}
\begin{table*}[!htbp]
    \ra{1.3}
    \centering
    %% local settings
    \sisetup{detect-weight,mode=text}
    % for avoiding siunitx using bold extended
    \renewrobustcmd{\bfseries}{\fontseries{b}\selectfont}
    \renewrobustcmd{\boldmath}{}
    % abbreviation
    \newrobustcmd{\B}{\bfseries}
    % shorten the intercolumn spaces
    %\addtolength{\tabcolsep}{-4.1pt}
    \scriptsize
    \setlength{\tabcolsep}{10pt}
    %\resizebox{\textwidth}{!}{
        \begin{tabular}{@{}c l l c c c c c c @{}}
            \toprule
            Dataset & Method & Supervision & EPE3D$\downarrow$ & Acc3D(0.05)$\uparrow$ & Acc3D(0.1)$\uparrow$ & Outliers3D$\downarrow$ & EPE2D$\downarrow$ & Acc2D$\uparrow$ \\
            \midrule
            & ICP~\cite{besl1992method} & - & 0.4062 & 0.1614 & 0.3038 & 0.8796 & 23.2280 & 0.2913\\
            & \textbf{Ours} & \textbf{Self-supervised} & \B 0.1696 & \B 0.2532 & \B 0.5501 & \B 0.8046 & \B 9.0234 & \B 0.3931 \\
            \cmidrule{2-9}
            & FlowNet3D~\cite{liu2019flownet3d} & Hybrid & 0.1136 & 0.4125 & 0.7706 & 0.6016 & 5.9740 & 0.5692\\
            FlyingThings3D & SPLATFlowNet~\cite{su2018splatnet} & Full & 0.1205 & 0.4197 & 0.7180 & 0.6187 & 6.9759 & 0.5512\\
            & Original BCL~\cite{gu2019hplflownet} & Full & 0.1111 & 0.4279 & 0.7551 & 0.6054 & 6.3027 & 0.5669\\
            & HPLFlowNet~\cite{gu2019hplflownet} & Full & 0.0804 & 0.6144 &  0.8555 & 0.4287 & 4.6723 & 0.6764\\
            & \textbf{Ours} & \textbf{Hybrid} & \B 0.0688 & \B 0.6703 & \B 0.8792 & \B 0.4036 & \B 4.1646 & \B 0.7019\\
            \cmidrule{1-9}
            & ICP~\cite{besl1992method} & - & 0.5181 & 0.0669 & 0.1667 & 0.8712 & 27.6752 & 0.1056\\
            & \textbf{Ours} & \textbf{Self-supervised} & \B 0.4154 & \B 0.2209 & \B 0.3721 & \B 0.8096 & \B 15.0605 & \B 0.3162\\
            \cmidrule{2-9}
            & FlowNet3D~\cite{liu2019flownet3d} & Hybrid & 0.1767 & 0.3738 & 0.6677 & 0.5271 & 7.2141 & 0.5093\\
            KITTI & SPLATFlowNet~\cite{su2018splatnet} & Full & 0.1988 & 0.2174 & 0.5391 & 0.6575 & 8.2306 & 0.4189\\
            & Original BCL~\cite{gu2019hplflownet} & Full & 0.1729 & 0.2516 & 0.6011 & 0.6215 & 7.3476 & 0.4411\\
            & HPLFlowNet~\cite{gu2019hplflownet} & Full & 0.1169 & 0.4783 & 0.7776 & 0.4103 & 4.8055 &  0.5938\\
            & \textbf{Ours} & \textbf{Hybrid} & \B 0.1034 & \B 0.4884 & \B 0.8224 & \B 0.3939 & \B 4.1278 & \B 0.6330 \\
            \bottomrule
        \end{tabular}
    %}
    \vspace{-6pt}
    \caption[Quantitative comparison to other methods]{
        \textbf{Quantitative comparison to other methods evaluated on FlyingThings3D~\cite{mayer2016large} and KITTI~\cite{Menze2015CVPR}.} Column \textit{Supervision} describes model's degree of supervision, differentiating between self-supervised, fully-supervised and hybrid training i.e. supervised with self-supervisory signals. The best results for a combination of a metric and a dataset are shown in bold.
    }
    \label{table:compareother}
\end{table*}
In this section, we present the results of our model trained and evaluated on FlyingThings3D~\cite{mayer2016large}. 
We evaluate the same model on KITTI~\cite{Menze2015CVPR} to test for generalization. 
We compare our model to state-of-the-art supervised methods in Table~\ref{table:compareother}.
We further conduct a qualitative evaluation and a comparison on both datasets.
We then perform an ablation study on the network components.
Finally, we present an error distribution analysis and compare our runtime efficiency to other methods.

\boldparagraph{Implementation Details.} We set the rotation weight $w_{rot} = 10$ to balance the magnitudes of translation and rotation outputs as proposed in \cite{kendall2015posenet, melekhov2017relative}.
We set the number of pose refinement iterations $K = 5$ following \cite{yuan2018iterative}.
Similarly to \cite{gu2019hplflownet}, we train our models using Adam~\cite{Kingma-Ba-ICLR-2015} with the initial learning rate $\alpha=10^{-4}$ for 85 epochs and multiply $\alpha$ by 0.7 every 35 subsequent epochs.

\boldparagraph{Evaluation Metrics.}
We adopt the scene flow evaluation metrics from \cite{gu2019hplflownet} and the camera pose metrics from \cite{shavit2019introduction, gao20206d}. 
We use the following evaluation metrics:
\begin{itemize}[itemsep=1pt,topsep=0pt,leftmargin=*]
    \item \textit{EPE3D} [m]. Mean of Euclidean distances between the predicted and ground truth pair of 3D vectors over all points.
    \item \textit{Acc3D(0.05)}. Strict notion of accuracy. Percentage of points with an EPE3D $<$ 0.05m or relative error $<$ 5\%. 
    \item \textit{Acc3D(0.1)}. Relaxed notion of accuracy. Percentage of points with EPE3D $<$ 0.1m or relative error $<$ 10\%. 
    \item \textit{Outliers3D}. Percentage of points with EPE3D $>$ 0.3m or relative error $>$ 10\%.
    \item \textit{EPE2D} [px]. Mean Euclidean distances between scene flow and ground truth projected into 2D over all points. 
    \item \textit{Acc2D}. Percentage of points with EPE2D $<$ 3px or relative error $<$ 5\%.
    \item \textit{RLE} [m]. Mean Euclidean distances between the predicted and ground-truth relative translations.
    \item \textit{ROE} [degrees]. Mean minimum rotation angles needed for aligning estimated and ground truth rotations.
\end{itemize}
\begin{figure*}[!htbp]
    \centering
    \scriptsize
    \setlength{\tabcolsep}{1pt}
    \newcommand{\sz}{0.24}
    \begin{tabular}{ccccc}
        \multirow{1}{*}[35pt]{\rotatebox{90}{Hybrid}} &
        \includegraphics[width=\sz\textwidth]{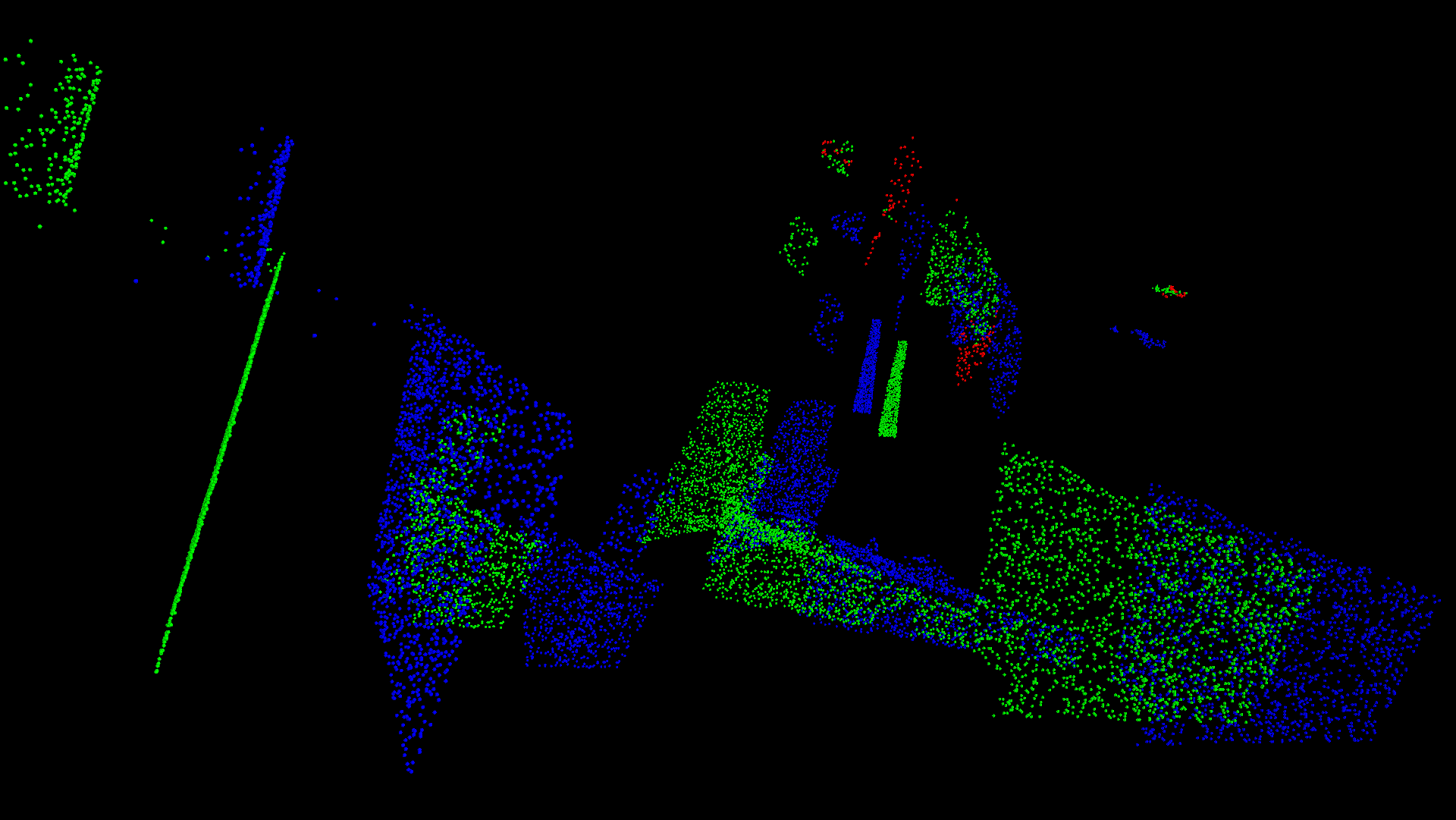} &
        \includegraphics[width=\sz\textwidth]{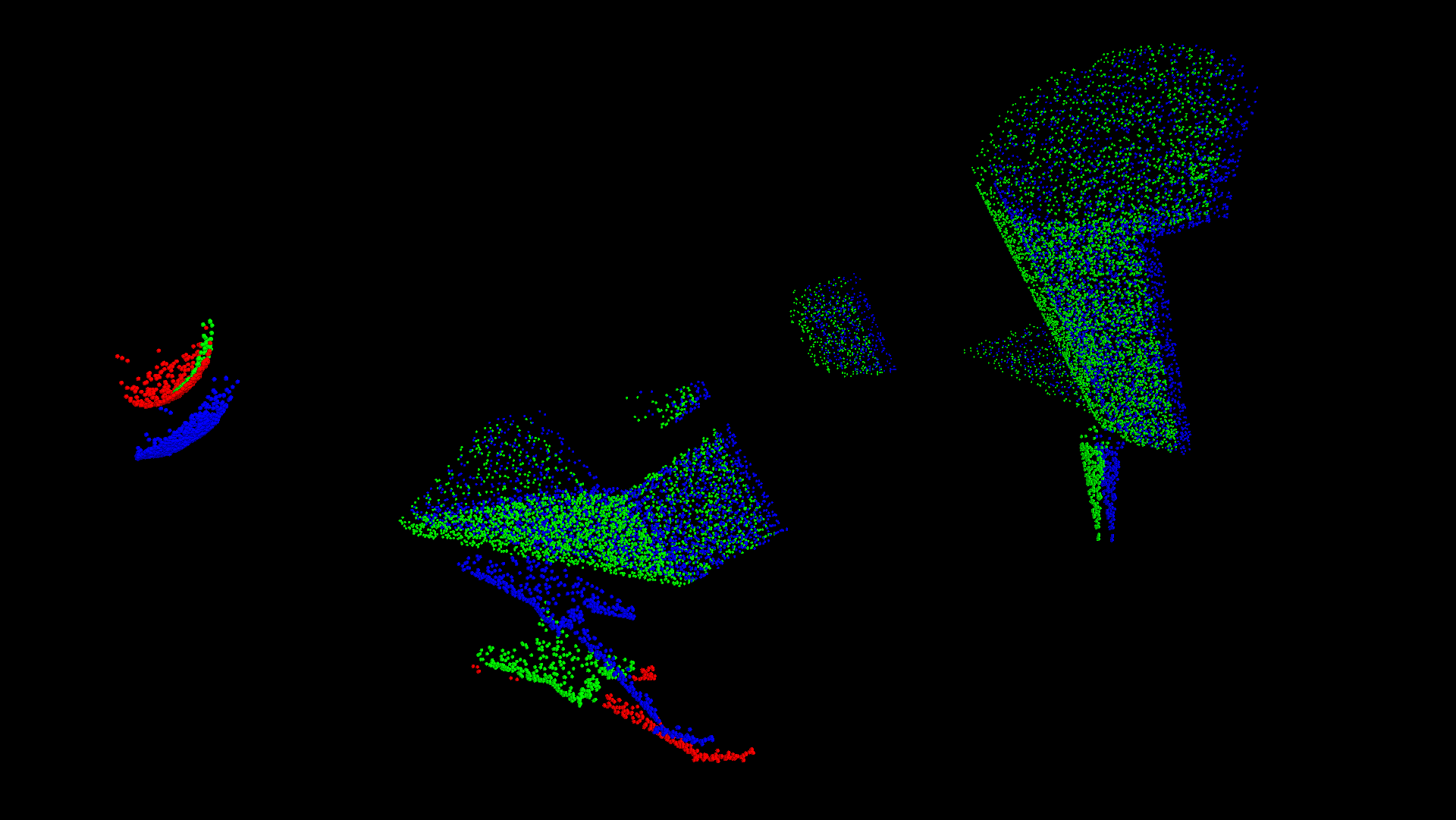} &
        \includegraphics[width=\sz\textwidth]{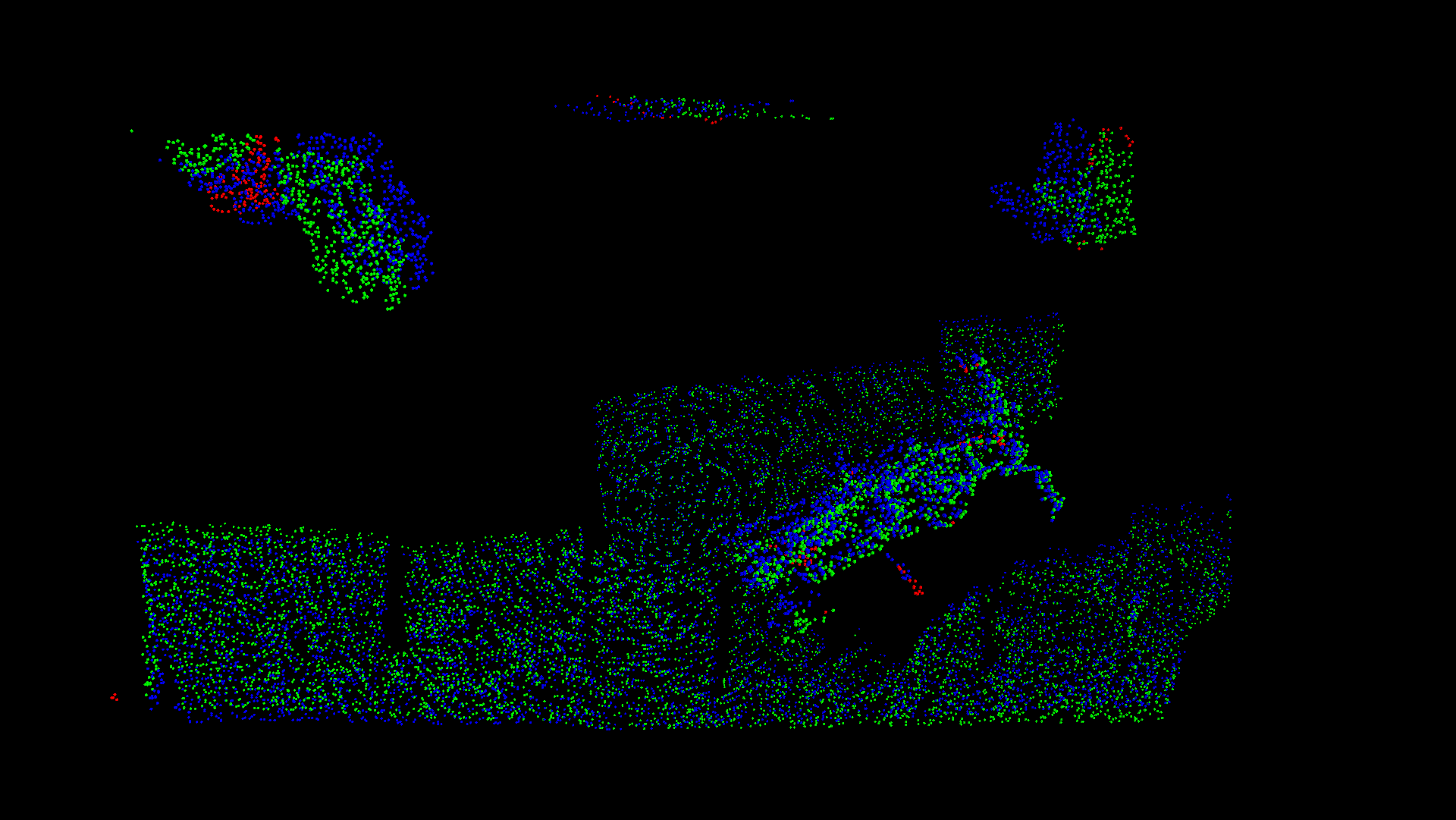} &
        \includegraphics[width=\sz\textwidth]{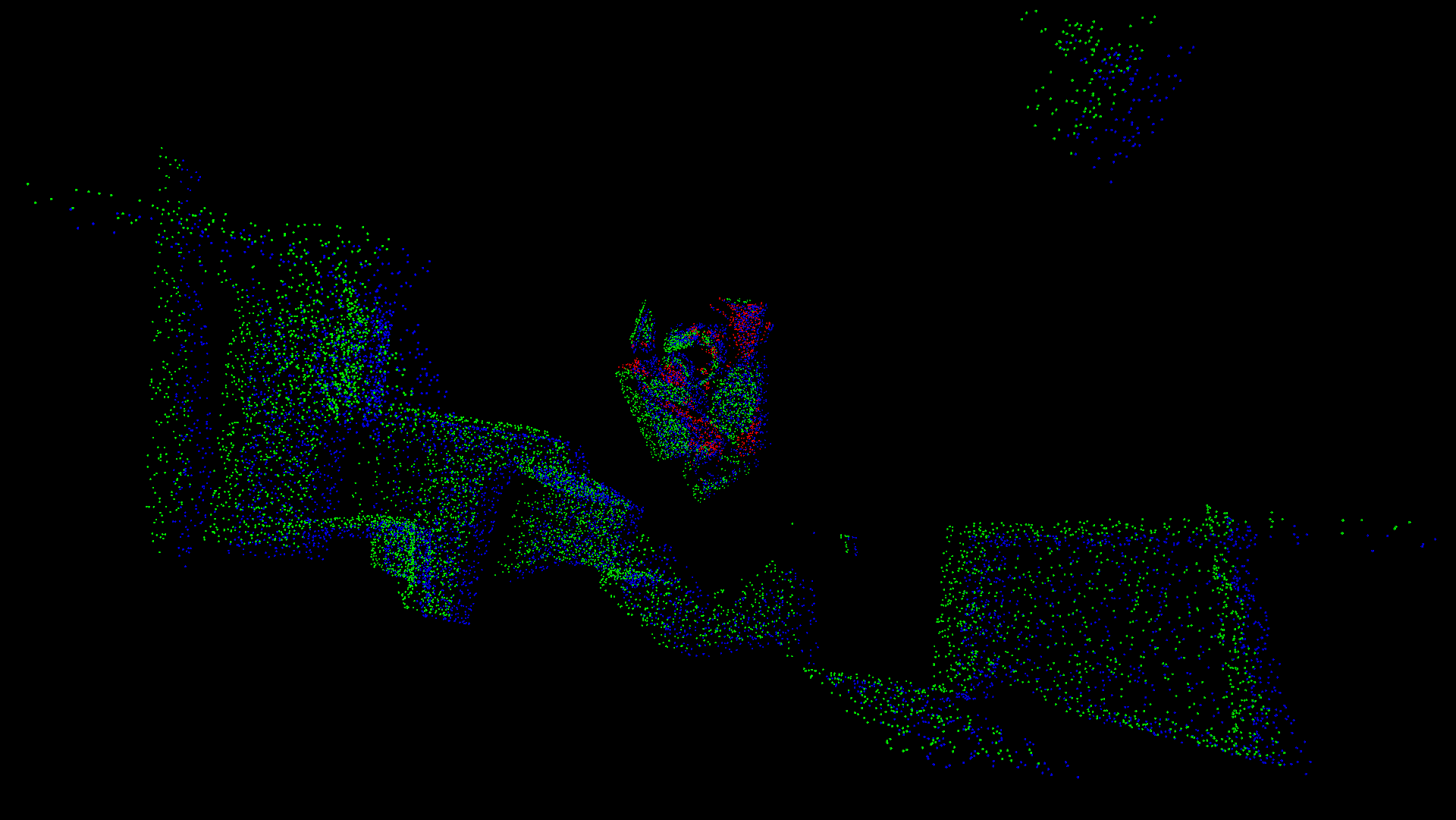} \\
        \multirow{1}{*}[50pt]{\rotatebox{90}{Self-Supervised}} &
        \includegraphics[width=\sz\textwidth]{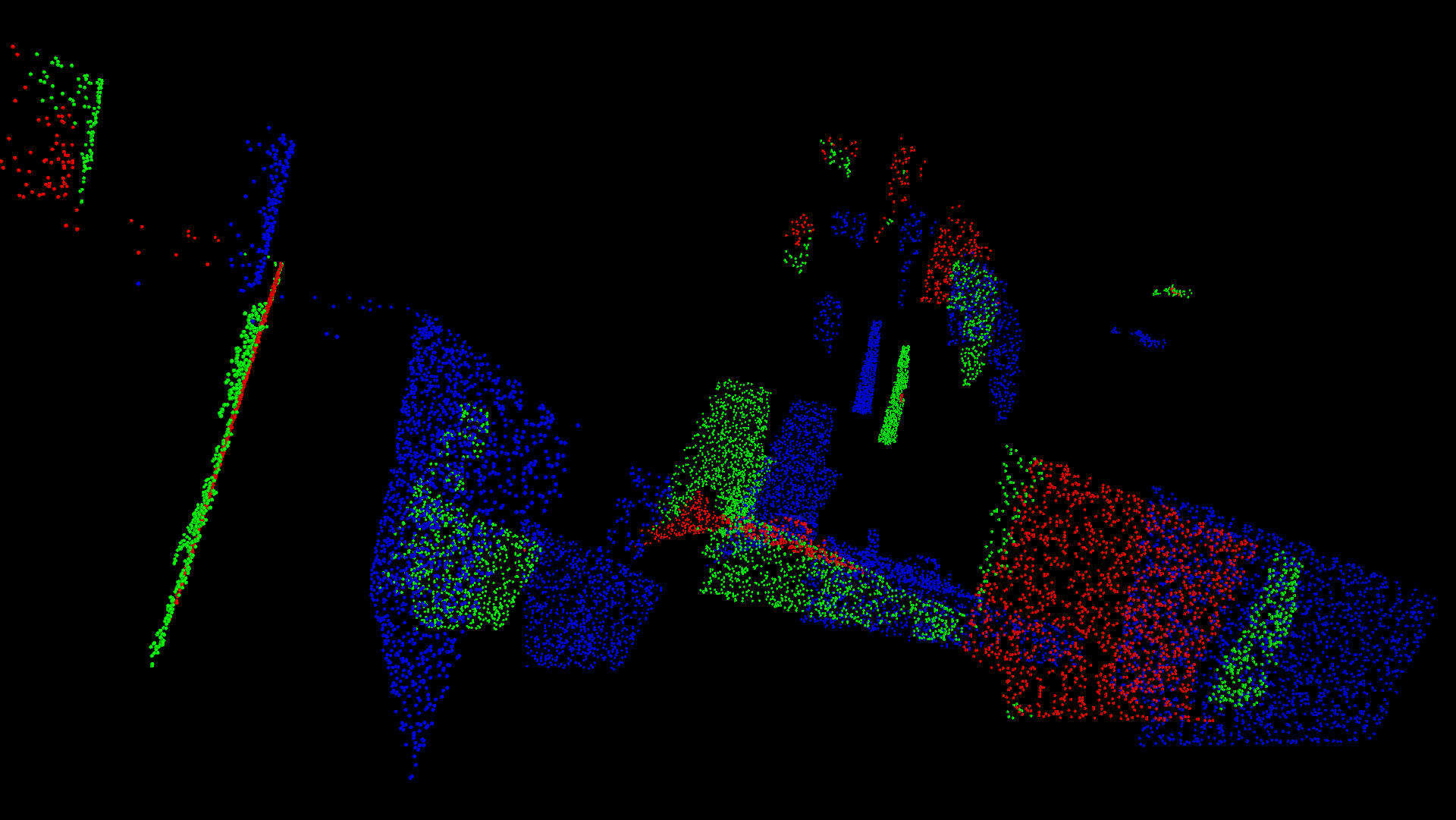} &
        \includegraphics[width=\sz\textwidth]{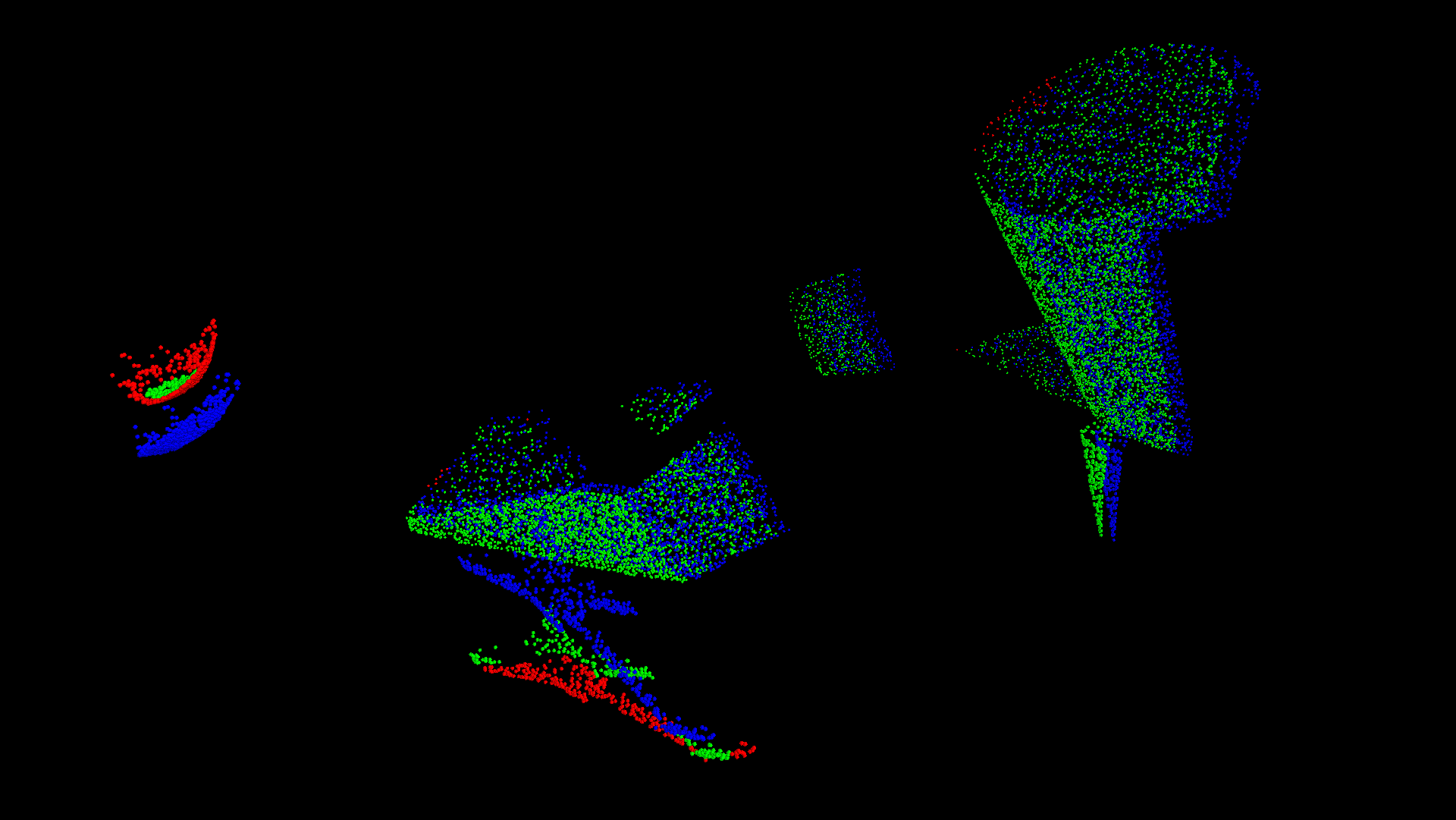} &
        \includegraphics[width=\sz\textwidth]{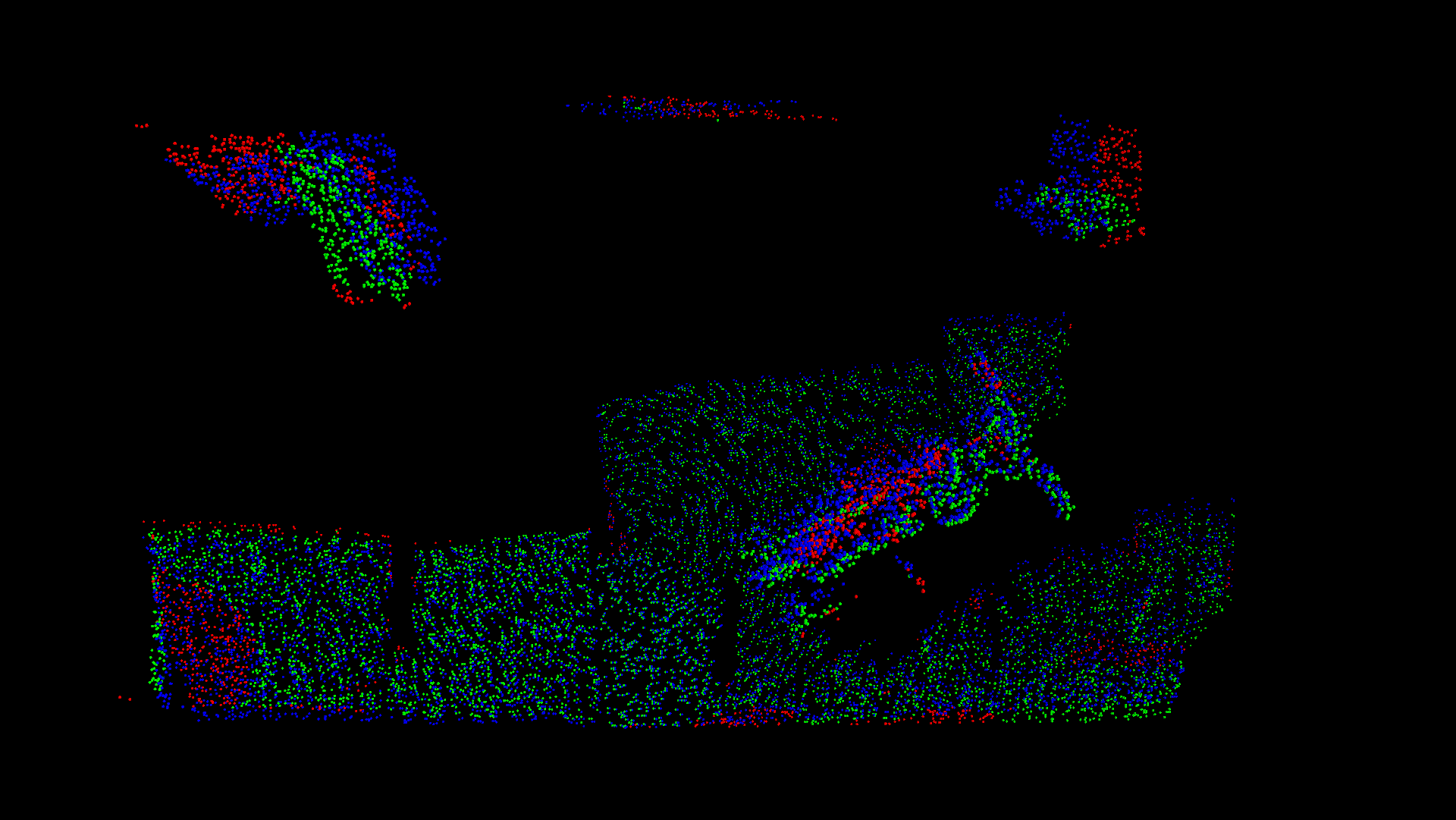} &
        \includegraphics[width=\sz\textwidth]{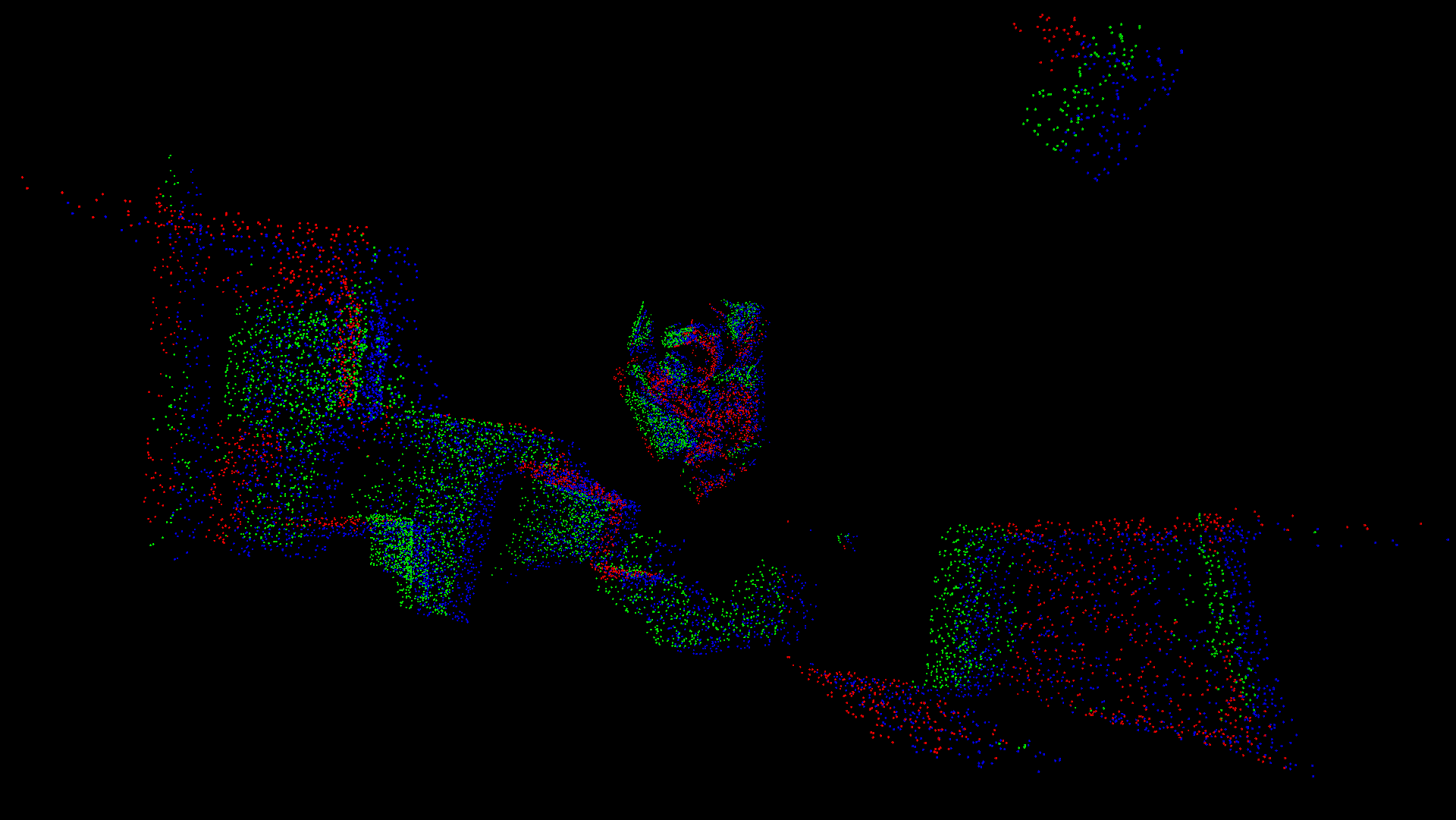} \\
        \multicolumn{5}{c}{FlyingThings3D} \\[+0.1cm]
        \multirow{1}{*}[35pt]{\rotatebox{90}{Hybrid}} &
        \includegraphics[width=\sz\textwidth]{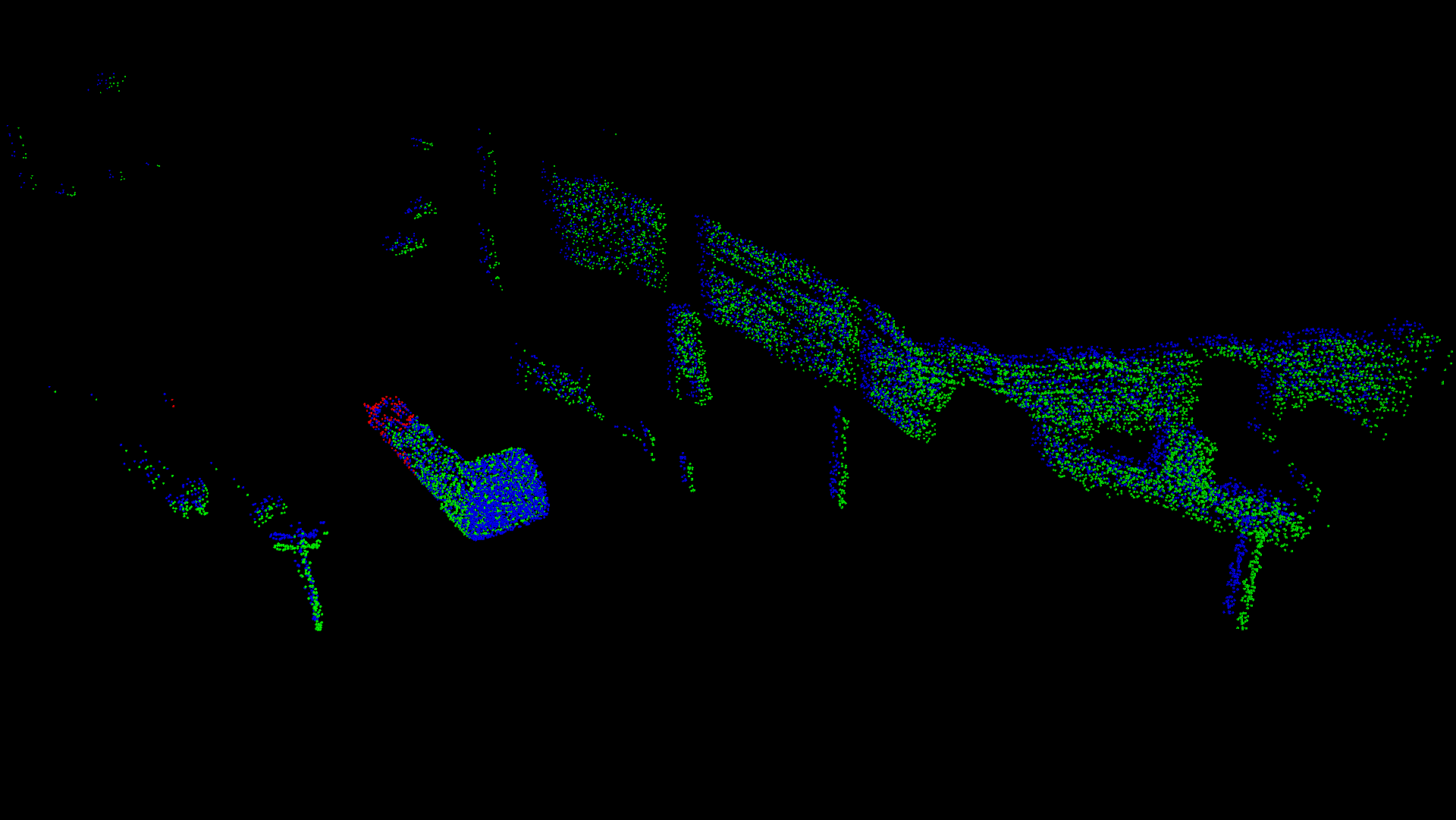} &
        \includegraphics[width=\sz\textwidth]{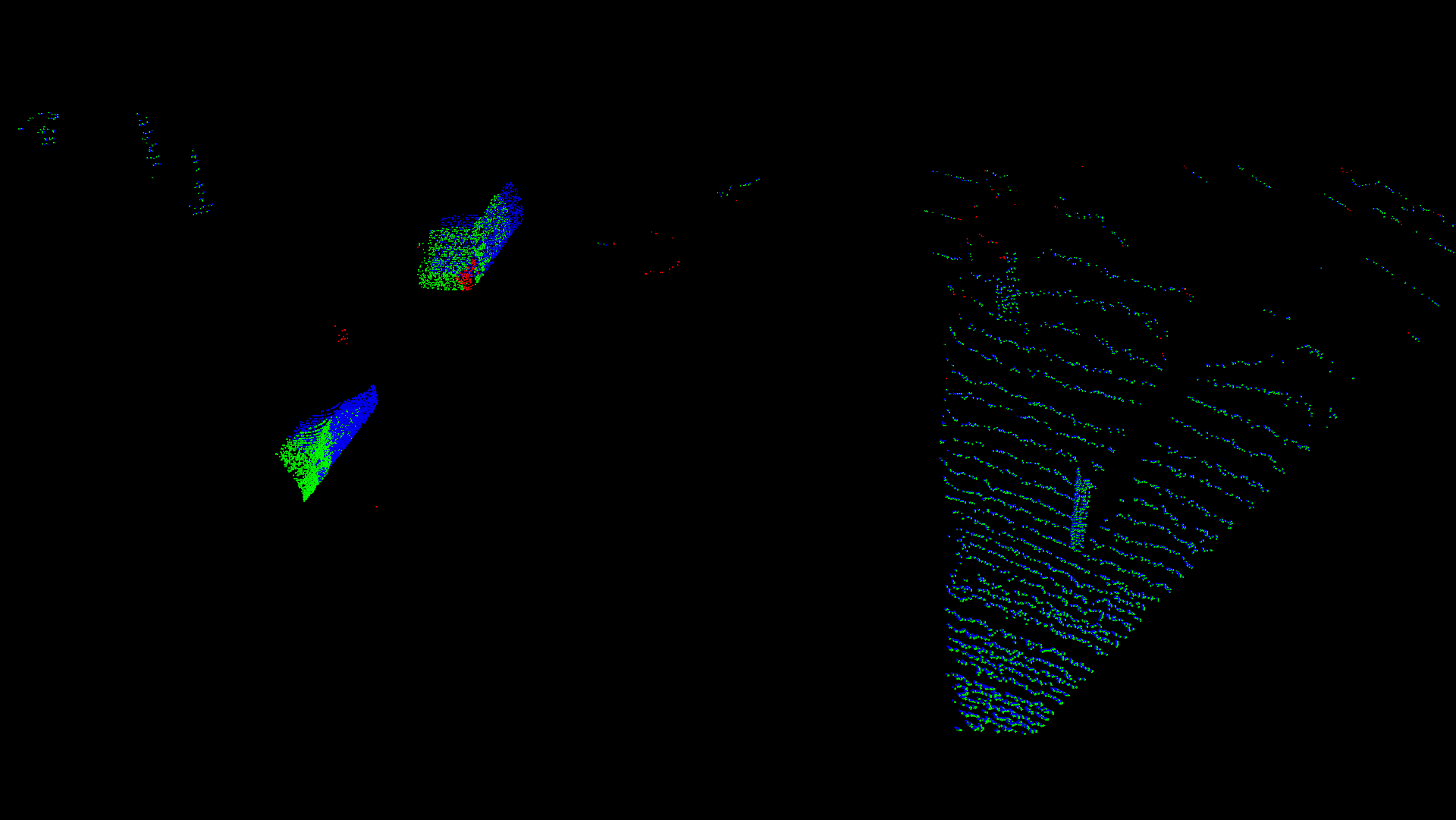} &
        \includegraphics[width=\sz\textwidth]{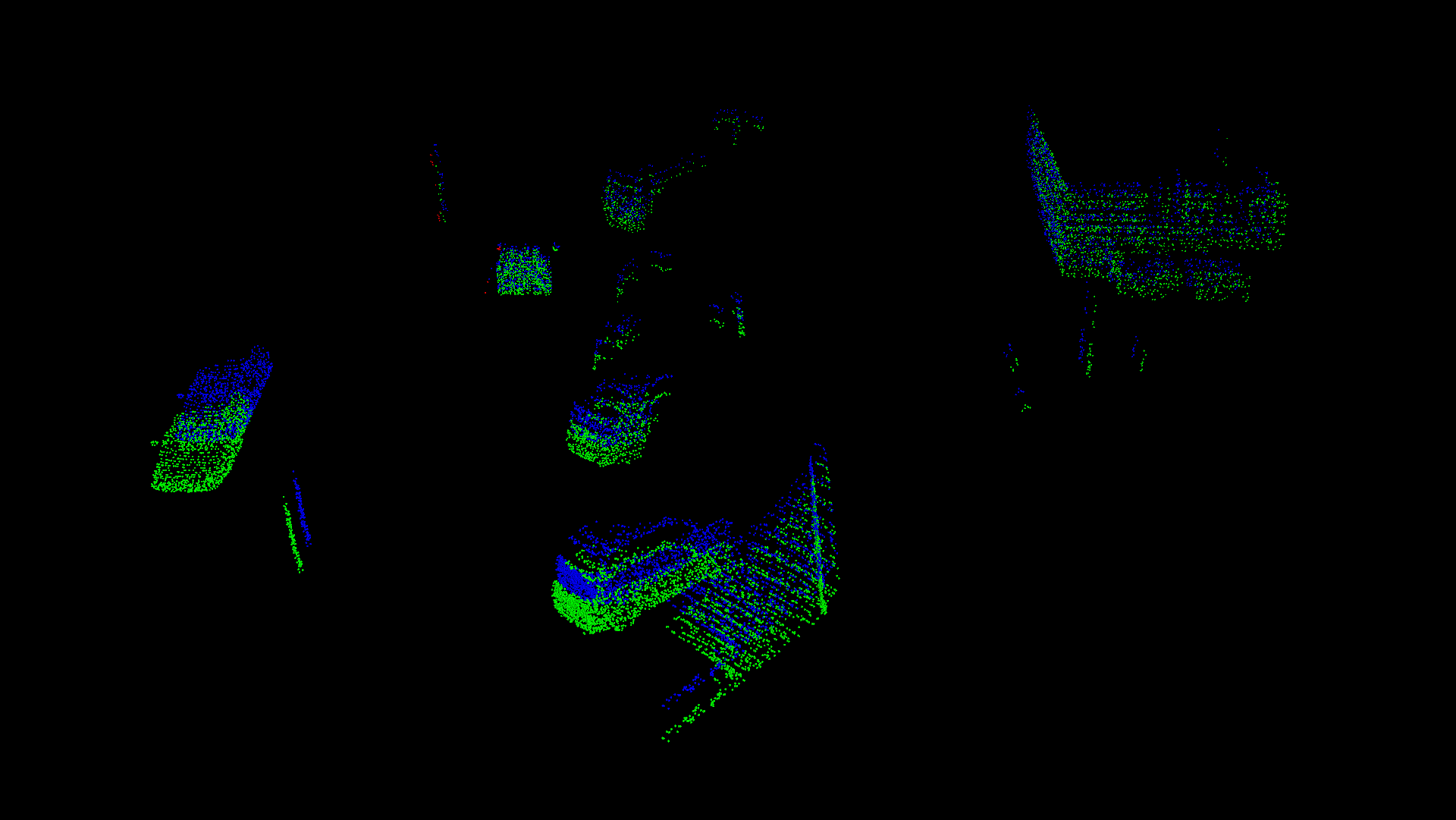} &
        \includegraphics[width=\sz\textwidth]{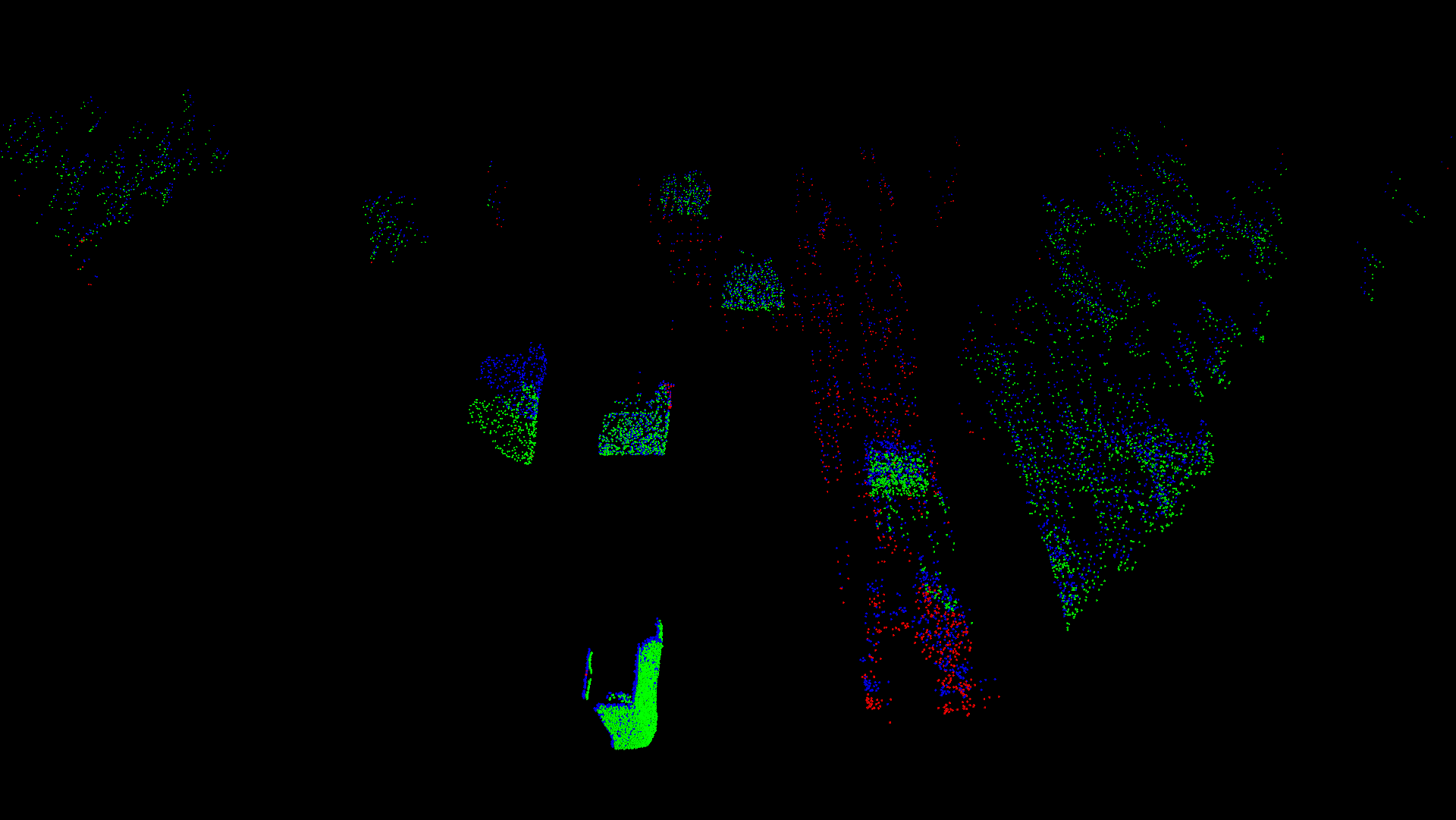} \\
        \multirow{1}{*}[50pt]{\rotatebox{90}{Self-Supervised}} &
        \includegraphics[width=\sz\textwidth]{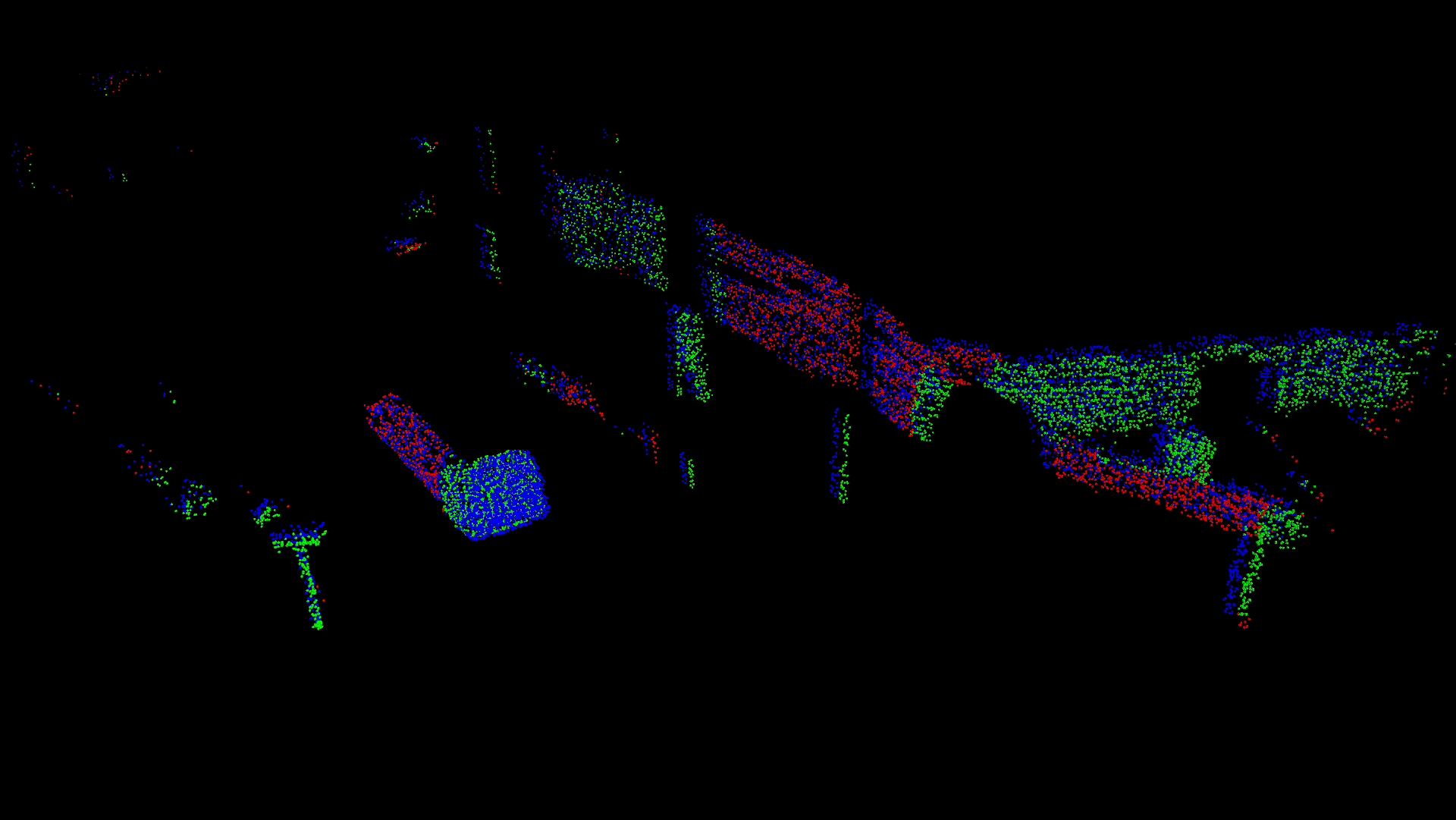} &
        \includegraphics[width=\sz\textwidth]{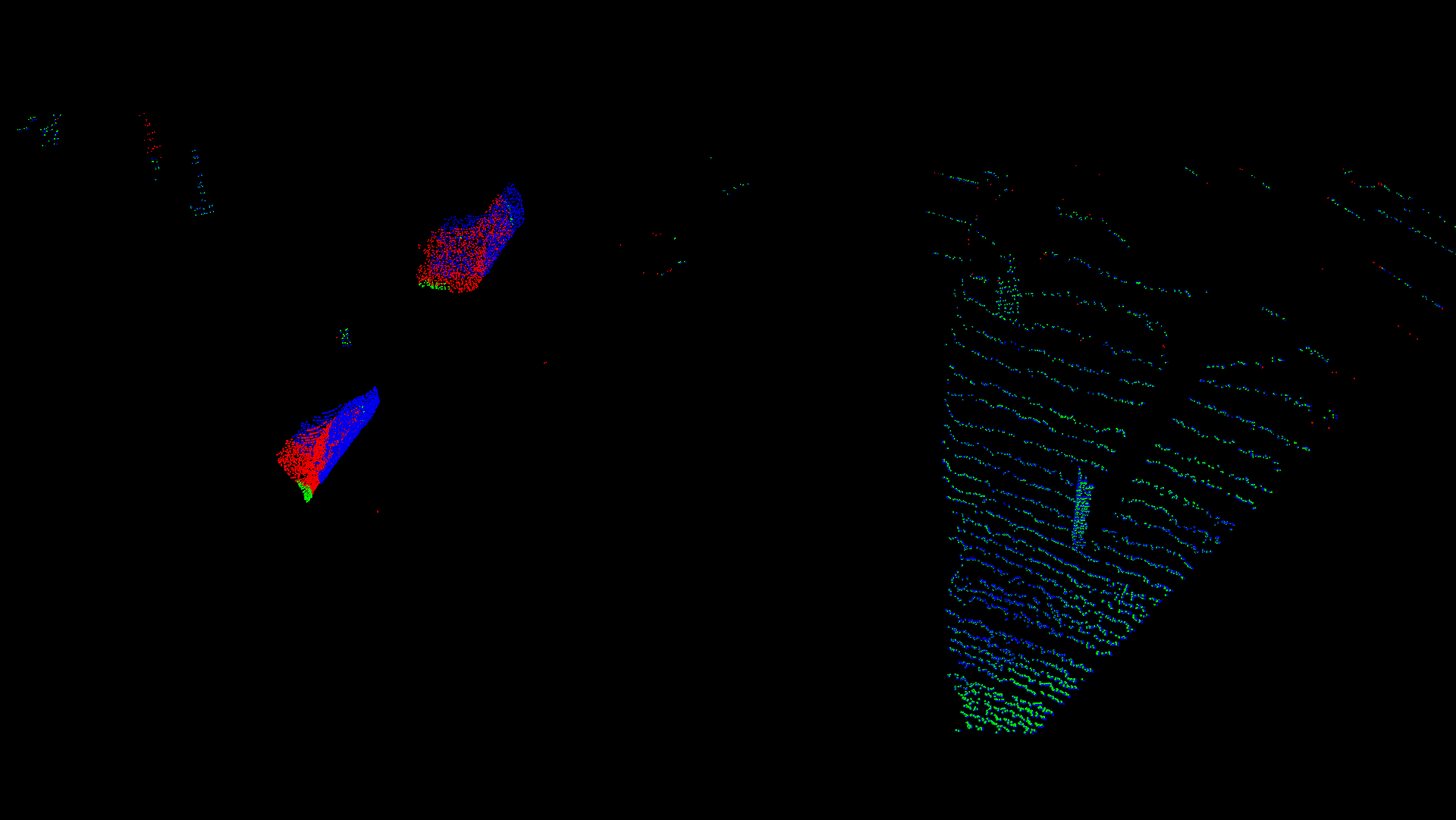} &
        \includegraphics[width=\sz\textwidth]{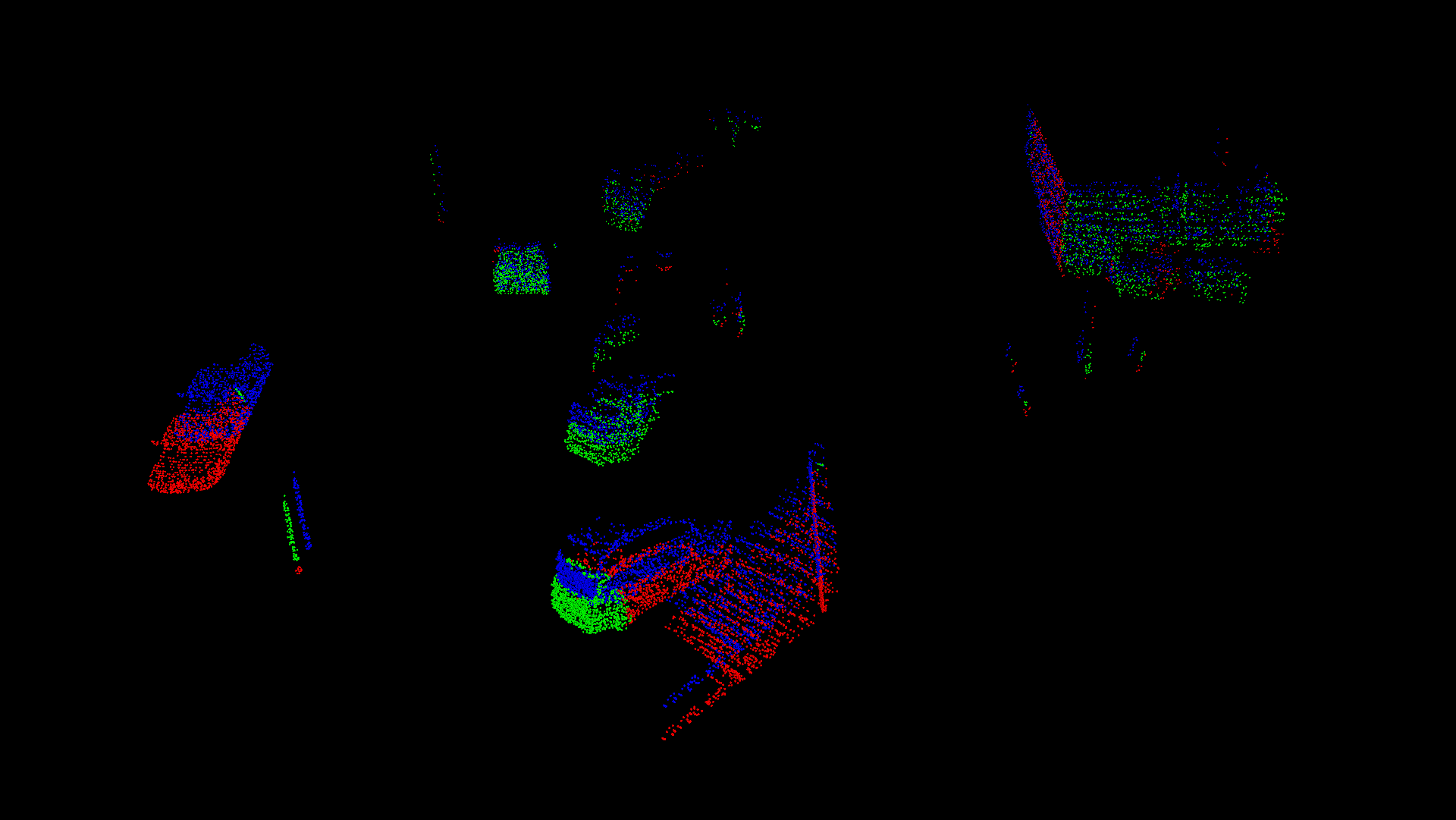} &
        \includegraphics[width=\sz\textwidth]{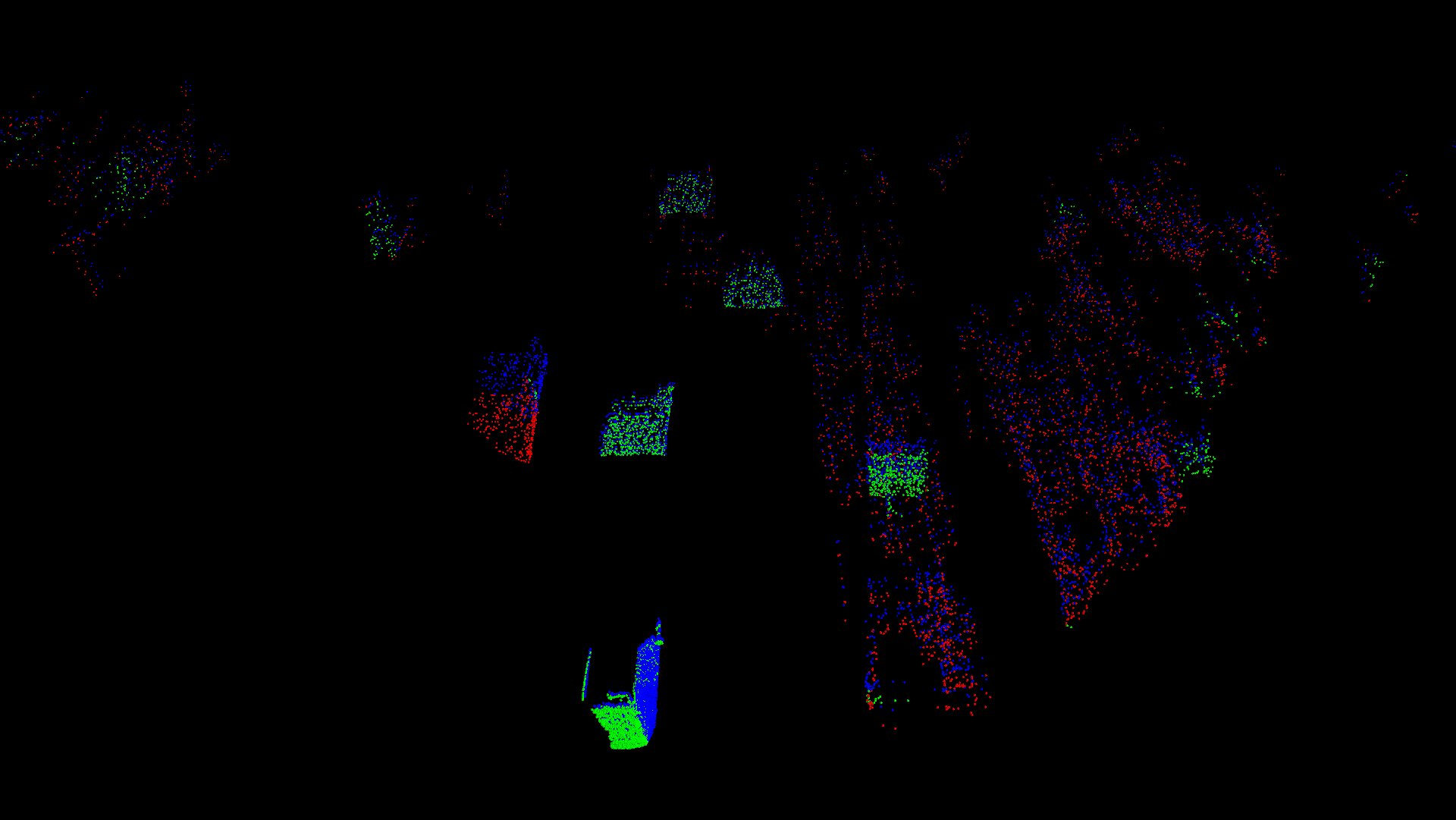} \\
        \multicolumn{5}{c}{KITTI} \\[-0.25cm]
    \end{tabular}
\caption{\textbf{Qualitative results on FlyingThings3D~\cite{mayer2016large} and KITTI~\cite{Menze2015CVPR}.} For each dataset the top row depicts the results from hybrid training (supervised training with self-supervisory terms). The bottom row shows the results from fully self-supervised training. The original input point cloud $\pc{1}$ is displayed in blue. Correctly predicted points are shown in green as a warped point cloud with predicted total flows $\pc{1} + \hat{D}$. Wrongly predicted points are shown in red as points warped with the ground-truth total flows $\pc{1} + {D}$. Correctness is defined according to Acc3D(0.1). Hybrid training showed better scene flow results according to smaller density of red regions. The self-supervised version demonstrated comparable overall results on FlyingThings3D but struggled on incomplete shapes and large motion in some scenes on KITTI.}
\label{fig:quality_fly_kitti}
\end{figure*}
\begin{figure*}[!htbp]
    \centering
    \scriptsize
    \setlength{\tabcolsep}{1pt}
    \newcommand{\sz}{0.46}
    \begin{tabular}{ccc}
        \multirow{1}{*}[80pt]{\rotatebox{90}{FlyingThings3D}} &
        \includegraphics[width=\sz\textwidth]{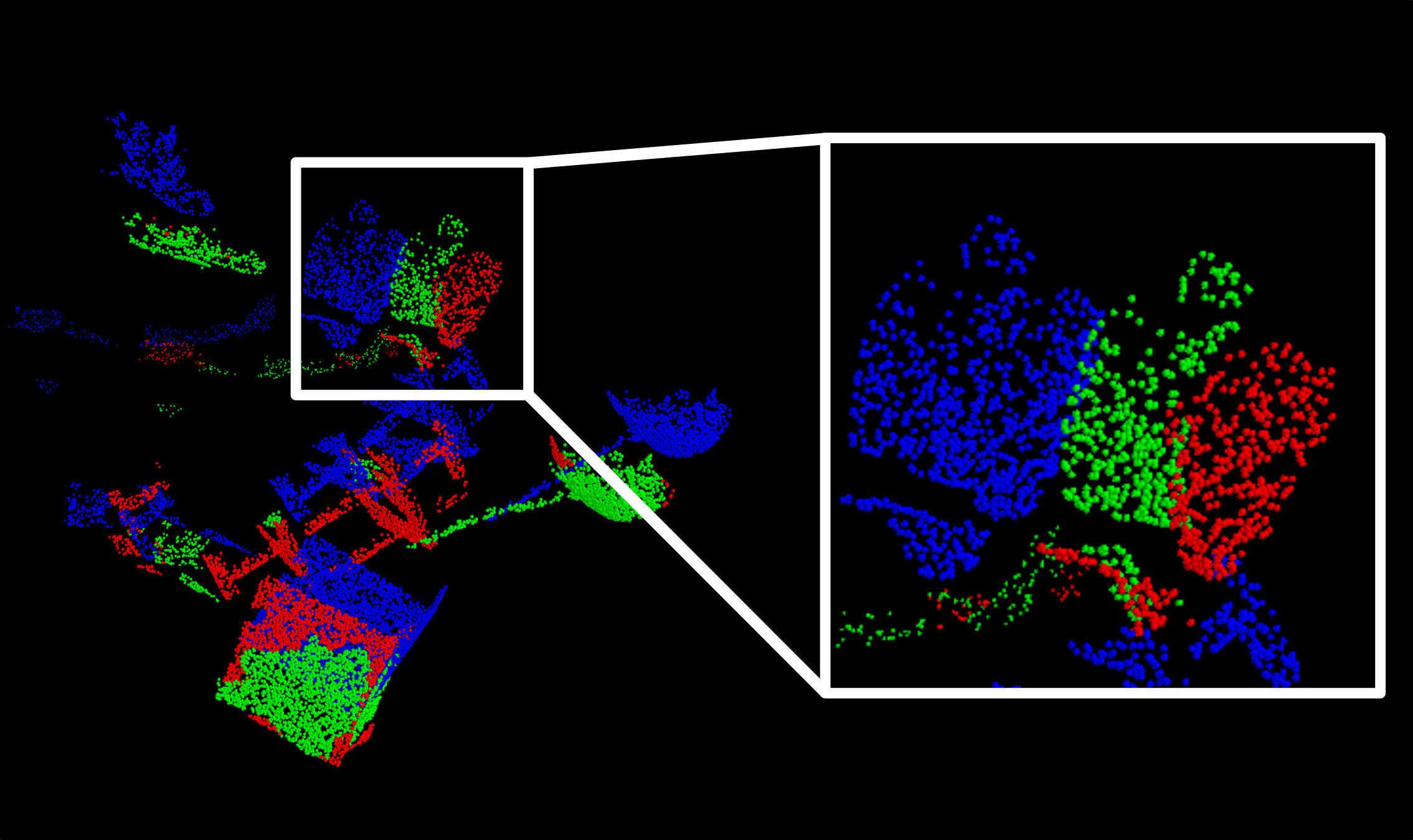} &
        \includegraphics[width=\sz\textwidth]{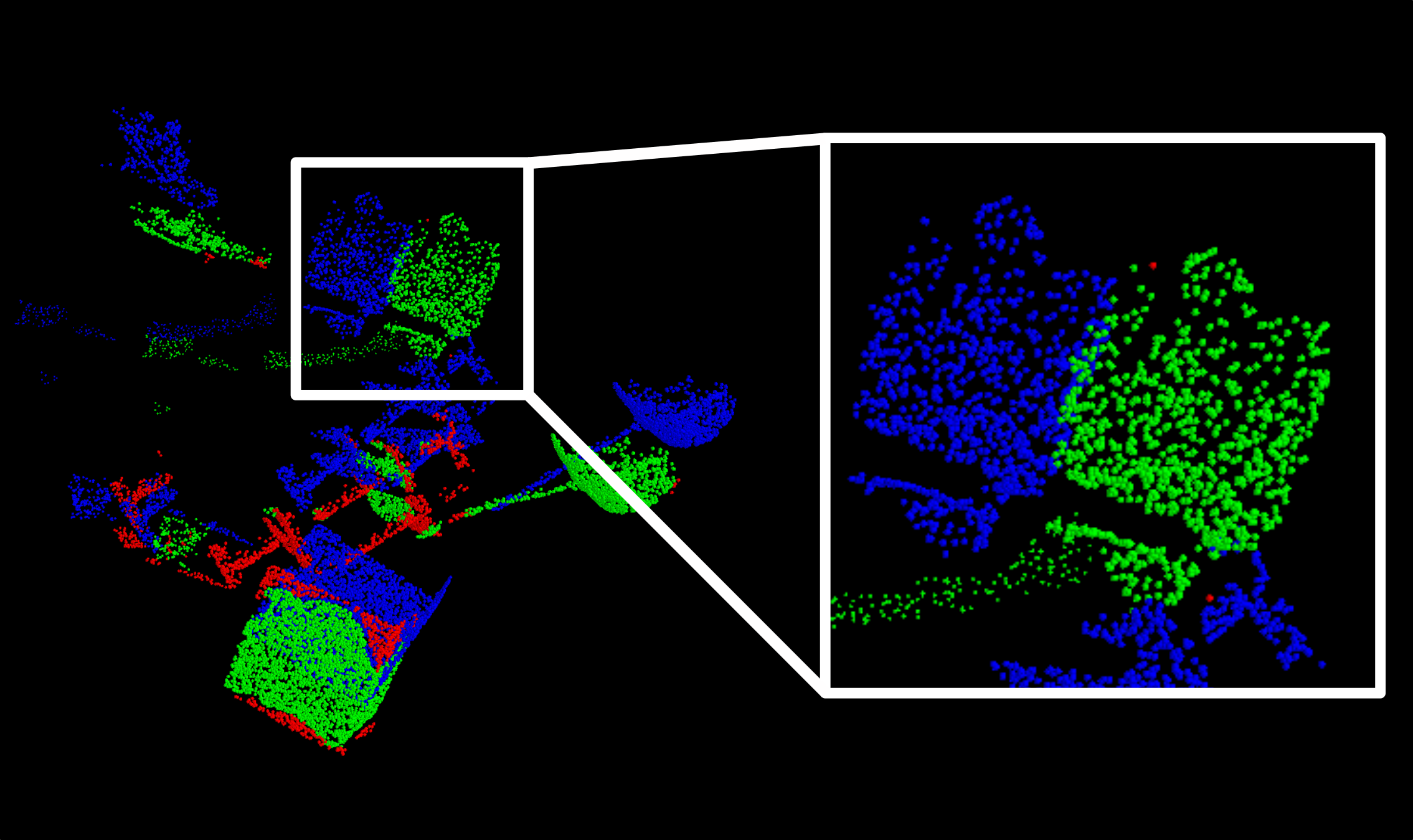} \\
        \multirow{1}{*}[70pt]{\rotatebox{90}{KITTI}} &
        \includegraphics[width=\sz\textwidth]{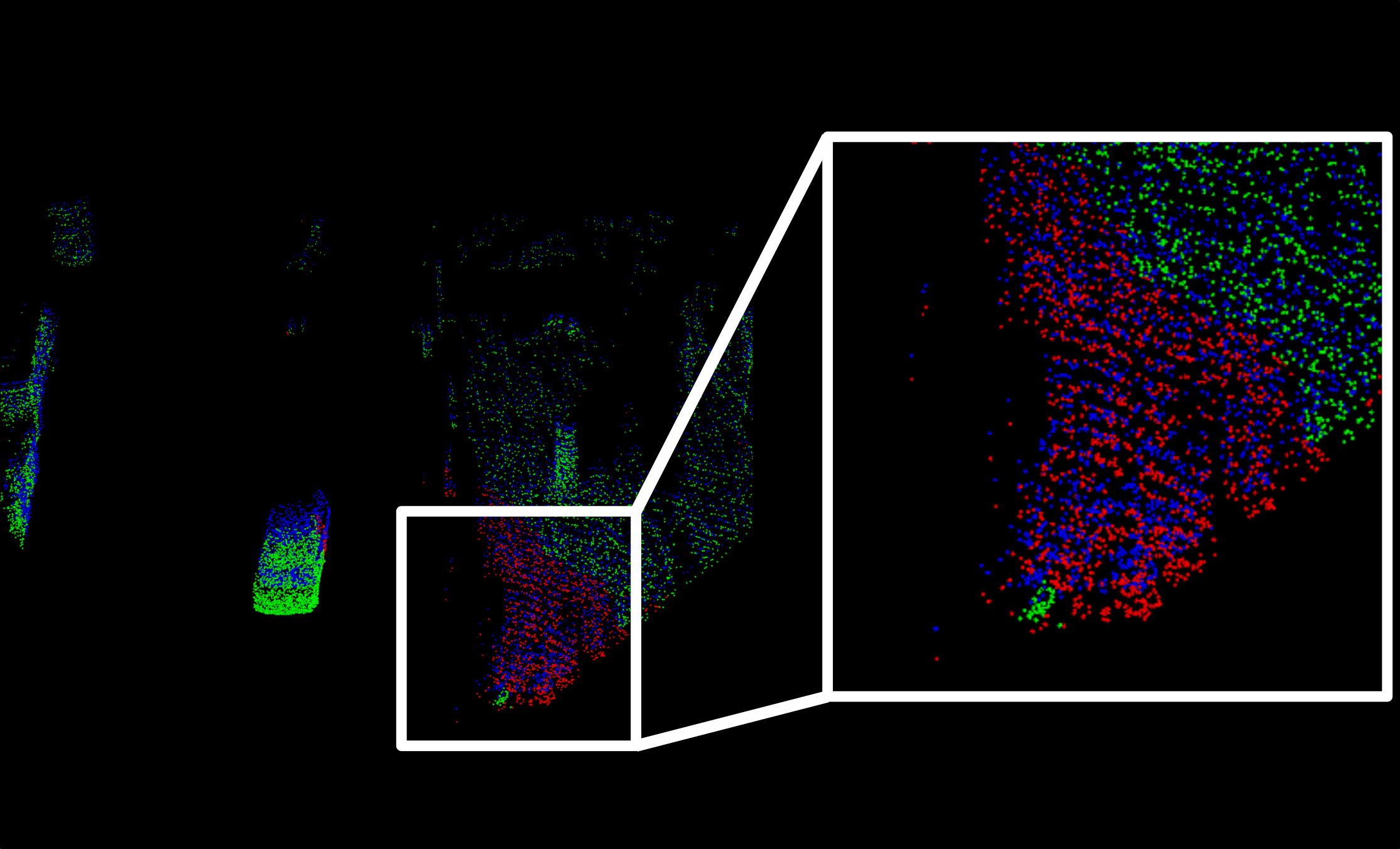} &
        \includegraphics[width=\sz\textwidth]{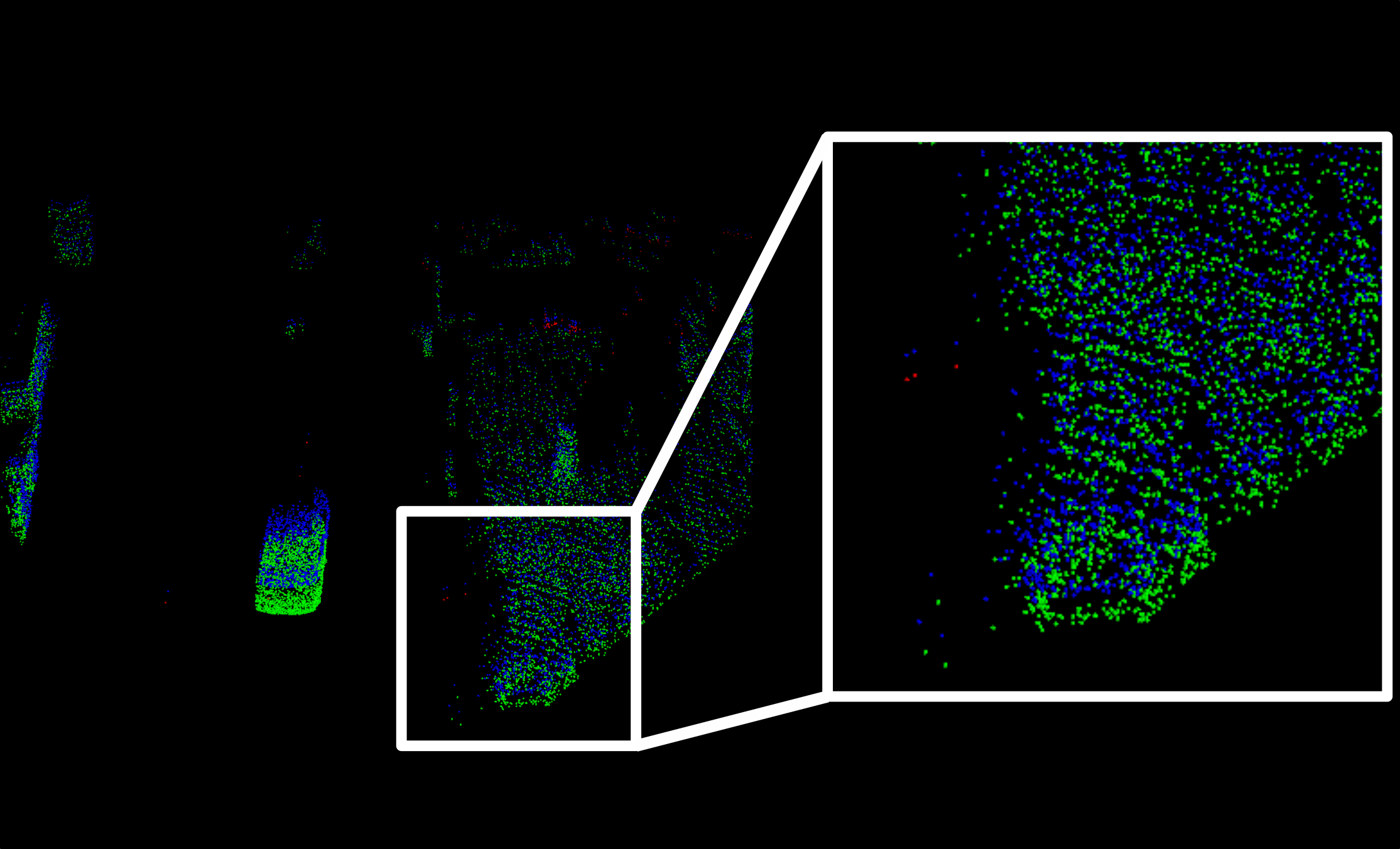} \\
        & HPLFlowNet~\cite{gu2019hplflownet} & Ours. Hybrid \\[-5pt]
    \end{tabular}
    \caption{\textbf{Qualitative comparison on FlyingThings3D~\cite{mayer2016large} and KITTI~\cite{Menze2015CVPR}.} The original input point cloud $\pc{1}$ is displayed in blue. Correctly predicted points are shown in green as a warped point cloud with predicted total flows $\pc{1} + \hat{D}$. Wrongly predicted points are shown in red as points warped with the ground-truth total flows $\pc{1} + {D}$. Our method generates fewer errors (red) on FlyingThings3D and shows better generalization ability on KITTI.}
    \label{fig:quality_baseline_fly_kitti}
    \vspace{-0.2cm}
\end{figure*}
\subsection{Results on FlyingThings3D}
\boldparagraph{Data Preprocessing.}
Similarly to \cite{gu2019hplflownet, wu2019pointpwc}, we use the \textsl{DispNet/FlowNet2.0} subset of FlyingThings3D~\cite{mayer2016large} where the extremely challenging pairs are removed.
Our method has the notation of ego-motion, therefore the ground truth for relative camera pose is needed to train the network.

Since the subset dataset of FlyingThings3D does not provide any relative pose ground truth, we match the frames of the subset dataset with the frames in the full dataset to obtain camera poses.
The ground truth for the camera poses of both FlyingThings3D uses the convention camera to world.
Therefore, the relative rotation is constructed using $R_{rel} = R_2^T R_1$ and relative translation is $t_{rel} = R_2^T (t_1 - t_2)$, which we combine to construct the ground-truth for the relative camera pose $T_{rel} = \begin{bmatrix}R_{rel} & t_{rel}\end{bmatrix}$. 

During matching of a subset and a full dataset we discovered that some frames are missing the ground-truth for the camera poses.
Consequently, we removed 64 observations with missing corresponding camera poses from the training/testing datasets which was used for the internal ablation study.
In contrast, when evaluating the results against other methods we use the full subset for a fair comparison.

The size of our training dataset is equal to 19586 pairs. We use 25\% of our training set which accounts to 4897 point cloud pairs in order to reduce the training times.
Following \cite{gu2019hplflownet} we evaluate our models on the whole testing set (3816 pairs).

\boldparagraph{Quantitative Results.}~We selected FlowNet3D~\cite{liu2019flownet3d}, SPLATFlowNet~\cite{su2018splatnet}, Original BCL~\cite{gu2019hplflownet} and HPLFlowNet~\cite{gu2019hplflownet} as baselines to compare our approach against the state-of-the-art supervised methods. From Table~\ref{table:compareother} we can deduce that our method outperformed the baselines consistently on all metrics. 
% The performance is expected to improve further given the same amount of training epochs as~\cite{gu2019hplflownet}.
%
For the self-supervised setting, we compare our method to ICP~\cite{besl1992method}.
As shown in Table~\ref{table:compareother}, our self-supervised model outperforms ICP by a large margin consistently on all metrics. This could be accredited to the refinement step which serves as an additional source of implicit self-supervision.
Corresponding qualitative results are shown in Figures~\ref{fig:quality_fly_kitti}, \ref{fig:quality_baseline_fly_kitti}.

\subsection{Generalization Results on KITTI}
\boldparagraph{Data Prepossessing.}~We evaluate our models on the KITTI Scene Flow Dataset~\cite{Menze2015ISA, Menze2015CVPR} in order to show their generalization ability to real-world datasets after training on FlyingThings3D without fine-tuning.
Following \cite{gu2019hplflownet} we remove the ground by height ($<$ 0.3 m) for an adequate comparison with other methods. The models are evaluated on all 142 point cloud pairs.

\boldparagraph{Quantitative Results.}~Table~\ref{table:compareother} shows that both of our methods were able to generalize to KITTI after training on FlyingThings3D without any fine-tuning. 
Our hybrid model outperformed all baselines on all metrics. Similarly, our self-supervised model distinctly outperformed ICP on all metrics.
%Our model trained in hybrid mode performed on par with HPLFlowNet in terms of EPE3D. Our self-supervised model distinctly outperformed ICP on all metrics. 
%
\subsection{Ablation Study}
We investigate the effect of the following two factors: \textbf{1.}~We study the performance gain after introducing a decomposed total scene flow model along with its corresponding architecture described in Section~\ref{sec:arch}. \textbf{2.}~We investigate the benefits of iterative pose refinement. Table~\ref{table:ablation} shows the results of an ablation study after training for 30 epochs.
\begin{table*}[!htbp]
    \ra{1.3}
    \centering
    \scriptsize
    \setlength{\tabcolsep}{7.5pt}
    % \begin{center}
        % \resizebox{\columnwidth}{!}{
    \begin{tabular}{@{}l c c c c c c c c c c@{}}
    \toprule
    Supervision & Ego-motion & Refinement & EPE3D $\downarrow$ & Acc3D(0.05) $\uparrow$ & Acc3D(0.1) $\uparrow$ & Outliers3D $\downarrow$ & EPE2D $\downarrow$ & Acc2D $\uparrow$ & ROE $\downarrow$ & RLE $\downarrow$\\
    \midrule
     & \checkmark & \checkmark & \textbf{0.1178} & \textbf{0.3766} & \textbf{0.7030} & \textbf{0.6729} & \textbf{7.0885} & \textbf{0.5391} & \textbf{0.6739} & \textbf{0.2113}\\
    \textbf{Hybrid} & \checkmark &  & 0.1352 & 0.2329 & 0.6099 & 0.7825 & 7.8843 & 0.4107 & 1.0149 & 0.2892\\
     &  &  & 0.1510 & 0.2111 & 0.5725 & 0.8132 & 8.2190 & 0.4006 & - & -\\
    \cmidrule{1-11}
     & \checkmark & \checkmark & \textbf{0.1305} & \textbf{0.3545} & \textbf{0.6451} & \textbf{0.7003} & \textbf{7.7212} & \textbf{0.5101} & \textbf{0.6501} & \textbf{0.2061}\\
    Full & \checkmark & & 0.2105 & 0.0856 & 0.3428 & 0.9353 & 11.6666 & 0.2273 & 1.0706 & 0.2923\\
     &  & & 0.1738 & 0.1204 & 0.4607 & 0.8728 & 9.4731 & 0.3190 & - & -\\
    \cmidrule{1-11}
     & \checkmark & \checkmark & \textbf{0.1746} & \textbf{0.2403} & \textbf{0.5414} & \textbf{0.8022} & \textbf{9.4109} & \textbf{0.3976} & \textbf{1.9821} & \textbf{0.3640}\\
    Self-supervised & \checkmark &  & 0.2691 & 0.1082 & 0.3375 & 0.9299 & 13.3613 & 0.2072 & 2.4799 & 0.3658\\
     &  &  & 0.2680 & 0.1111 & 0.3394 & 0.9348 & 13.5647 & 0.2133 & - & -\\
    \bottomrule
    \end{tabular}
        % }
    % \end{center}
    \vspace{-6pt}
    \caption{\textbf{Ablation study on FlyingThigs3D~\cite{mayer2016large} for fully supervised, hybrid and self-supervised settings.} Hybrid mode training provided consistently better results on the majority of metrics. The decomposed total scene flow model (described in Section~\ref{sec:arch}) was able to improve the performance for the case of hybrid training. An iterative refinement with $K = 5$ has improved scene flow metrics and relative camera pose metrics alike on all configurations. ROE and RLE are not applicable in the case of total flow learning. All models are compared after training for 30 epochs.}
    \label{table:ablation}
    \vspace{-0.1cm}
\end{table*}
\begin{figure*}[!htbp]
    \centering
    \scriptsize
    \setlength{\tabcolsep}{15pt}
    \newcommand{\sz}{0.44}
    \begin{tabular}{cc}
        FlyingThings3D & KITTI \\[-1pt]
        \includegraphics[width=\sz\textwidth]{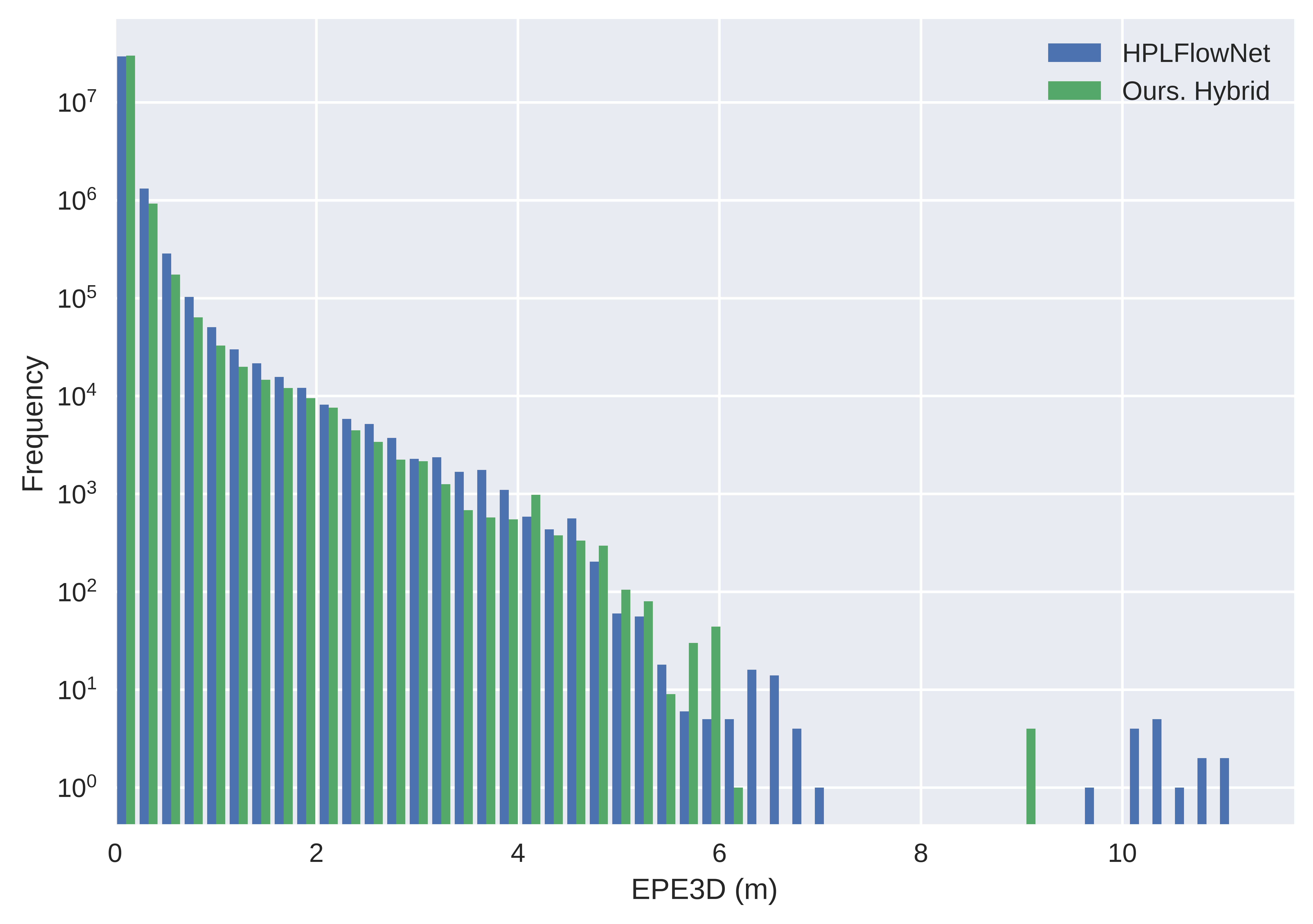} &
        \includegraphics[width=\sz\textwidth]{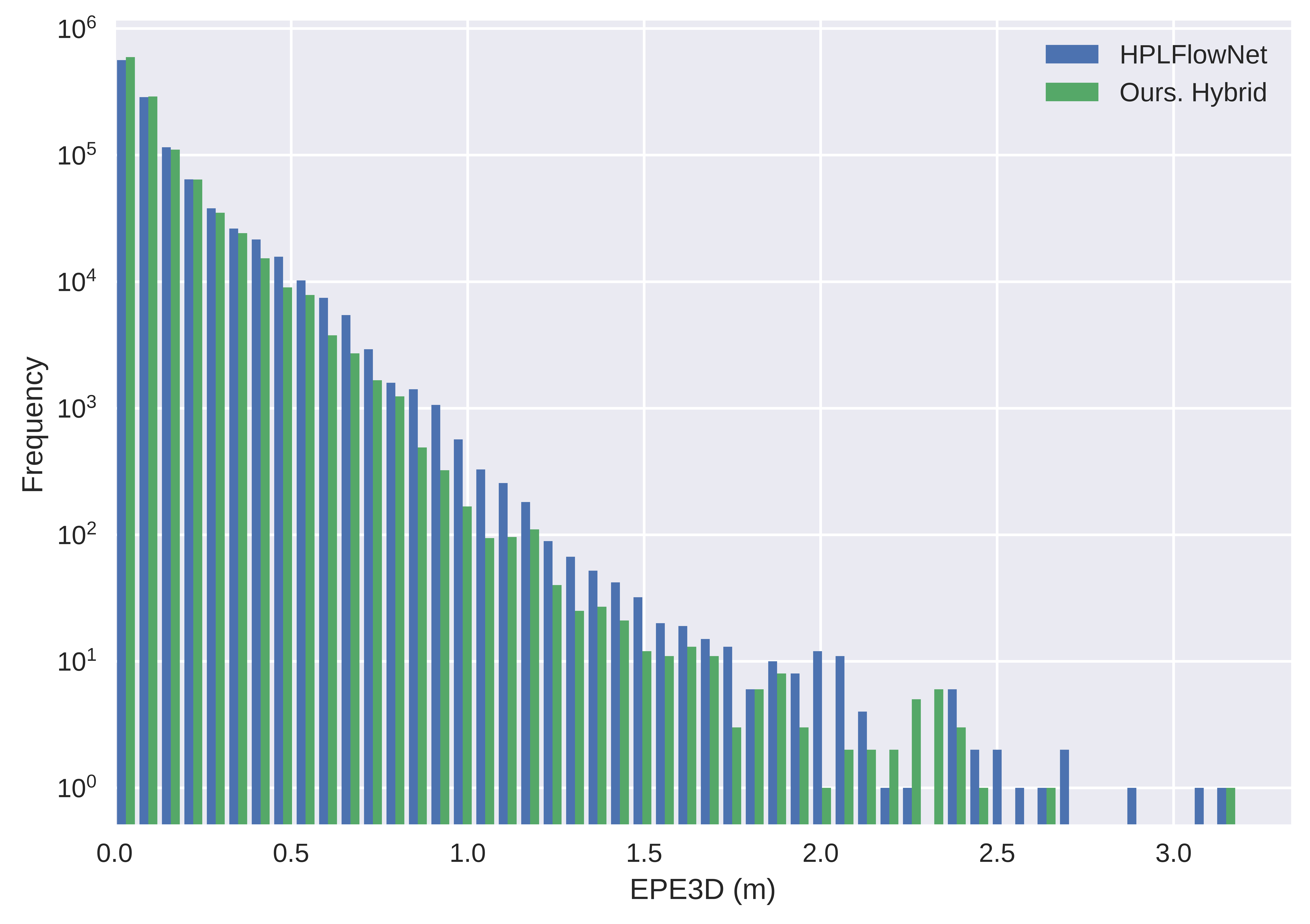} \\[-12pt]
    \end{tabular}
    \caption{\textbf{Distribution of End Point Error 3D (EPE3D) values on FlyingThings3D~\cite{mayer2016large} (left) and KITTI~\cite{Menze2015CVPR} (right).} We can deduce that our method is more robust to outliers since it contains significantly less outliers at the tail for both datasets. Our lower mean of EPE3D on FlyingThings3D can be accredited to the lower number of extreme outliers. Similarly, the amount of outliers on KITTI at the tail of the distribution is sparser for our method which explains the lower mean of EPE3D.}
    \label{fig:histo}
    \vspace{-0.1cm}
\end{figure*}
\subsection{Error Distribution Analysis}
In order to account for outliers, we compare the distributions of EPE3D values between our method and the baseline HPLFlowNet~\cite{gu2019hplflownet}. EPE3D values are computed for each point in every scene of the test sets for both FlyingThings3D~\cite{mayer2016large} and KITTI~\cite{Menze2015CVPR} and are distributed into 50 bins along the x-axis. We employ a logarithmic scale for the y-axis (frequencies) for visualization purposes to better demonstrate the outliers in the error distributions since the original curves reassemble an exponential decay. 
We present the histogram comparison of EPE3D in Figure~\ref{fig:histo}.
\begin{table}[!htb]
    \ra{1.3}
    \centering
    \scriptsize
    \setlength{\tabcolsep}{50pt} %20
    \begin{tabular}{@{}l r@{}}
        \toprule
        Method & Inference time [ms] $\downarrow$ \\
        \midrule
        FlowNet3D~\cite{liu2019flownet3d} & 130.8 \\
        HPLFlowNet~\cite{gu2019hplflownet} & 98.4 \\
        PointPWC-Net~\cite{wu2019pointpwc} & 117.4 \\
        Ours & 491.3 \\
        Ours + Refinement (K = 5) & 1539.1 \\
        \bottomrule
    \end{tabular}
    \vspace{-6pt}
    \caption{\textbf{Mean inference times on FlyingThings3D~\cite{mayer2016large}.} Note that all methods were run on different machines.}
    \label{table:time}
    \vspace{-0.3cm}
\end{table}
\subsection{Runtime Efficiency}
Table~\ref{table:time} shows mean inference times for major baselines to indicate the order of magnitude as they were run on different machines.
Average inference times for FlowNet3D~\cite{liu2019flownet3d}, HPLFlowNet~\cite{gu2019hplflownet}, PointPWC-Net~\cite{wu2019pointpwc} are adopted from \cite{gu2019hplflownet, wu2019pointpwc}.
Our model has higher runtime due to more weights and the recurrent nature of the pose regression network.
%Higher runtimes of our models are explained by higher number of weights and the recurrent nature of the pose regression network.
%------------------------------------------------------------------------
\section{Conclusion}
We presented an alternative approach for scene flow estimation which decomposes the total flow into ego-motion flow learned with a pose regression network and non-rigid flow learned with a non-rigid flow network.
Experiments performed on both FlyingThings3D and KITTI datasets demonstrated that scene flow decomposition and self-supervision can improve the performance of scene flow estimation for both hybrid and self-supervised modes.
Especially the hybrid training scheme was consistently better than any other training scheme.
Moreover, ablation studies showed that iterative refinement improves the performance even further at the cost of extra computation.

%------------------------------------------------------------------------
{
\boldparagraph{Acknowledgments.}
This work has been supported by Innosuisse funding (Grant No. 34475.1 IP-ICT).
}
%------------------------------------------------------------------------
%%%%%%%%% SUPPLEMENT
\clearpage
\appendix
%%%%%%%%% TITLE
\centerline{\large\bf Appendix}%
\vspace*{12pt}%
%%%%%%%%% BODY TEXT
% Overview 
%------------------------------------------------------------------------
In this appendix, we provide additional details and results for our method. 
We give further details on the network architecture and the training procedure in Section~\ref{sec:impl}. 
In Section~\ref{sec:data}, additional information on the data prepossessing and augmentations is provided. 
Section~\ref{sec:exp} contains an extended analysis of our experiments. Finally, we present additional qualitative results in Section~\ref{sec:quality}. 
Our source code will be made publicly available upon publication.

\section{Implementation Details} \label{sec:impl}
We used batch size $B = 1$ for all of our experiments. 
For hybrid training we set the weights for the corresponding loss terms to $w_{epe3d} = w_{nr} = w_{r} = 1$.
In case of fully self-supervised training we set the weights to $w_{epe3d} = w_{nr} = w_{r} = 0$, such that the total loss is only represented by the self-supervised loss $L = L_{ss}$. 
Our best performing hybrid model used in the quantitative comparison was trained for a total of $457$ epochs with training refinement iterations set to $K_t = 1$. 
Our best self-supervised model was trained for a total of $88$ epochs with additional training refinement iterations $K_t = 5$.

%------------------------------------------------------------------------
\section{Data Preprocessing} \label{sec:data}
We use the same data augmentation as in \cite{gu2019hplflownet} which consists of the following transformations: scaling, rotations, translations.
We apply the augmentations on the ground truth of both the point clouds as well the relative pose, although the non-rigid transformations (\eg scaling) have been omitted from the augmentation pipeline.

%------------------------------------------------------------------------
\section{Extended Experiments} \label{sec:exp}
\boldparagraph{Refinement Iterations Analysis.}
Due to the recurrent architecture of the relative pose regression network it is possible to trade higher runtimes for lower errors.
We examine inference refinement iterations $K_i \in \{1, 5, 10, 20, 30\}$ by doing the inference using the weights of our best performing self-supervised model trained with the number of training iterations $K_t = 5$. 
In Table~\ref{table:inference}, we elaborate on how the number of inference iterations $K_i$ affects the performance on the scene flow metrics during evaluation.
\begin{table*}[!htbp]
    \ra{1.3}
    \centering
    \sisetup{detect-weight,mode=text}
    \renewrobustcmd{\bfseries}{\fontseries{b}\selectfont}
    \renewrobustcmd{\boldmath}{}
    \newrobustcmd{\B}{\bfseries}
    \scriptsize
    \setlength{\tabcolsep}{12pt}
        \begin{tabular}{@{}c c c c c c c c r@{}}
            \toprule
            Dataset & $K_i$ & EPE3D$\downarrow$ & Acc3D(0.05)$\uparrow$ & Acc3D(0.1)$\uparrow$ & Outliers3D$\downarrow$ & EPE2D$\downarrow$ & Acc2D$\uparrow$ & Inference time [ms] $\downarrow$ \\
            \midrule
            & 1 & 0.2030 & 0.1990 & 0.4836 & 0.8580	& 10.5297 & 0.3432 & \textbf{542.8}\\
            & 5 & 0.1759 & 0.2415 & 0.5397 & 0.8014 & 9.5154 & 0.3979 & 1539.1 \\
            FlyingThings3D & 10 & 0.1714 & 0.2465 & 0.5456 & 0.7946	& 9.3642 & 0.4034 & 2715.0\\
            & 20 & 0.1689 & \textbf{0.2520} & \textbf{0.5481} & \textbf{0.7902} & \textbf{9.2688} & \textbf{0.4072} & 5231.0\\
            & 30 & \textbf{0.1687} & 0.2498 & 0.5475 & 0.7914 & 9.2783 & 0.4065 & 7853.8\\
            \cmidrule{1-9}
            & 1 & 0.4713 & 0.1688 & 0.3279 & 0.8518	& 16.5619 & 0.2707 & \textbf{485.8}\\
            & 5 & 0.4653 & 0.1647 & 0.3337 & 0.8547 & 16.3952 & 0.2711 & 1380.9\\
            KITTI & 10 & 0.4653	& \textbf{0.1716} & 0.3323 & 0.8535& 16.3442	& 0.2756 & 2474.0\\
            & 20 & 0.4595 & 0.1682 & \textbf{0.3375} & \textbf{0.8490} & 16.2655 & \textbf{0.2763} & 4700.8\\
            & 30 & \textbf{0.4585} & 0.1621 & 0.3328 & 0.8554 & \textbf{16.2457} & 0.2717 & 6930.0 \\
            \bottomrule
        \end{tabular}
    \caption{\textbf{Influence of inference iterations on the total scene flow evaluation metrics.} We take the model trained with the number of training iterations $K_{t} = 5$ in self-supervised mode and perform inference on it while varying the number of inference iterations $K_{i}$. Increasing the number of inference iterations $K_i$ helps to improve the metrics at the cost of longer runtimes. An increase of $K_i$ consistently helps to improve EPE3D during inference for both datasets. Similarly, large values of $K_i$ mostly help to improve the performance on other metrics.}
    \label{table:inference}
\end{table*}

%------------------------------------------------------------------------
\section{Additional Qualitative Evaluation} \label{sec:quality}
\begin{figure*}[!htbp]
    \centering
    \scriptsize
    \setlength{\tabcolsep}{1pt}
    \newcommand{\sz}{0.24}
    \begin{tabular}{ccccc}
        \multirow{1}{*}[35pt]{\rotatebox{90}{Hybrid}} &
        \includegraphics[width=\sz\textwidth]{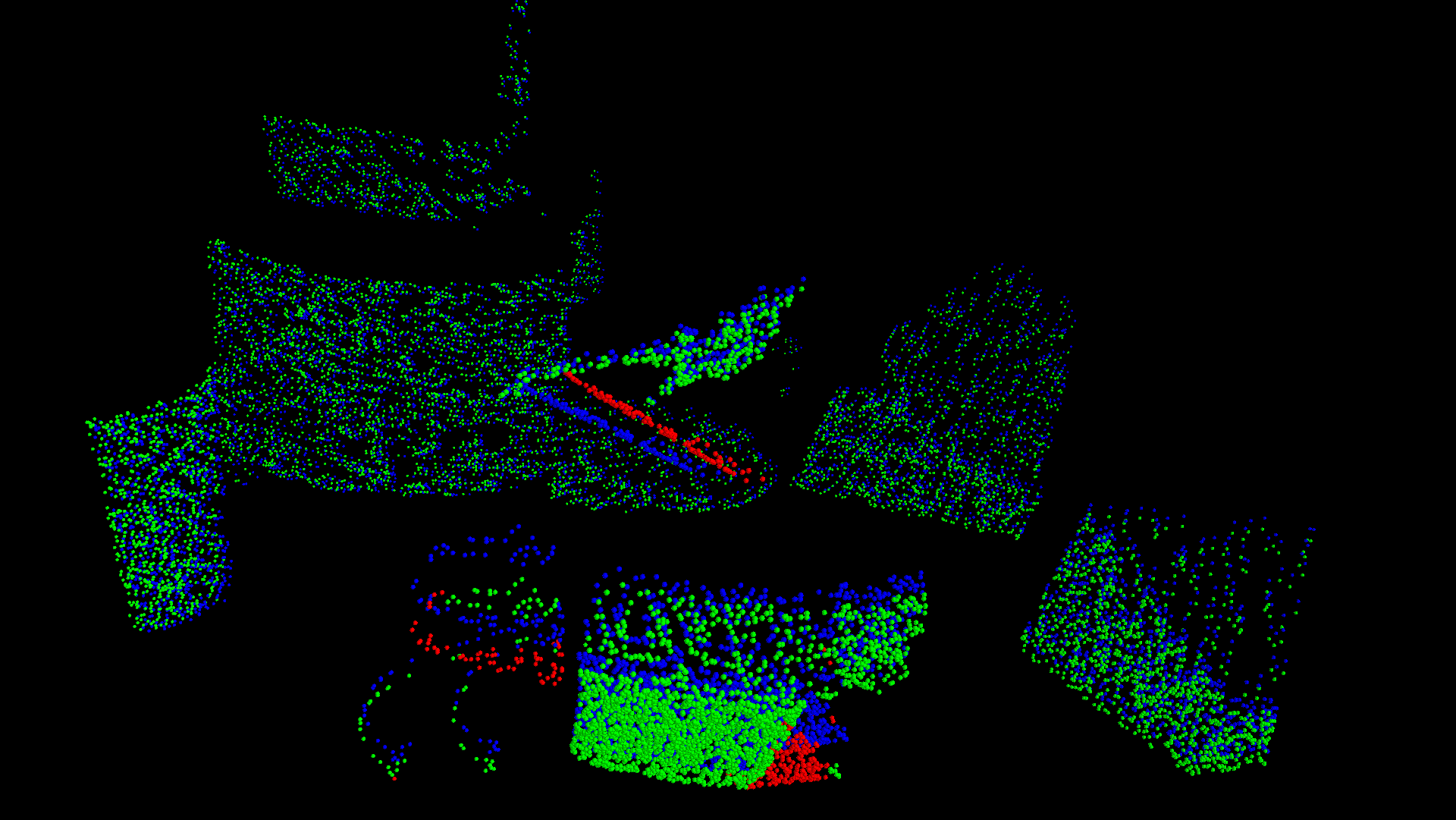} &
        \includegraphics[width=\sz\textwidth]{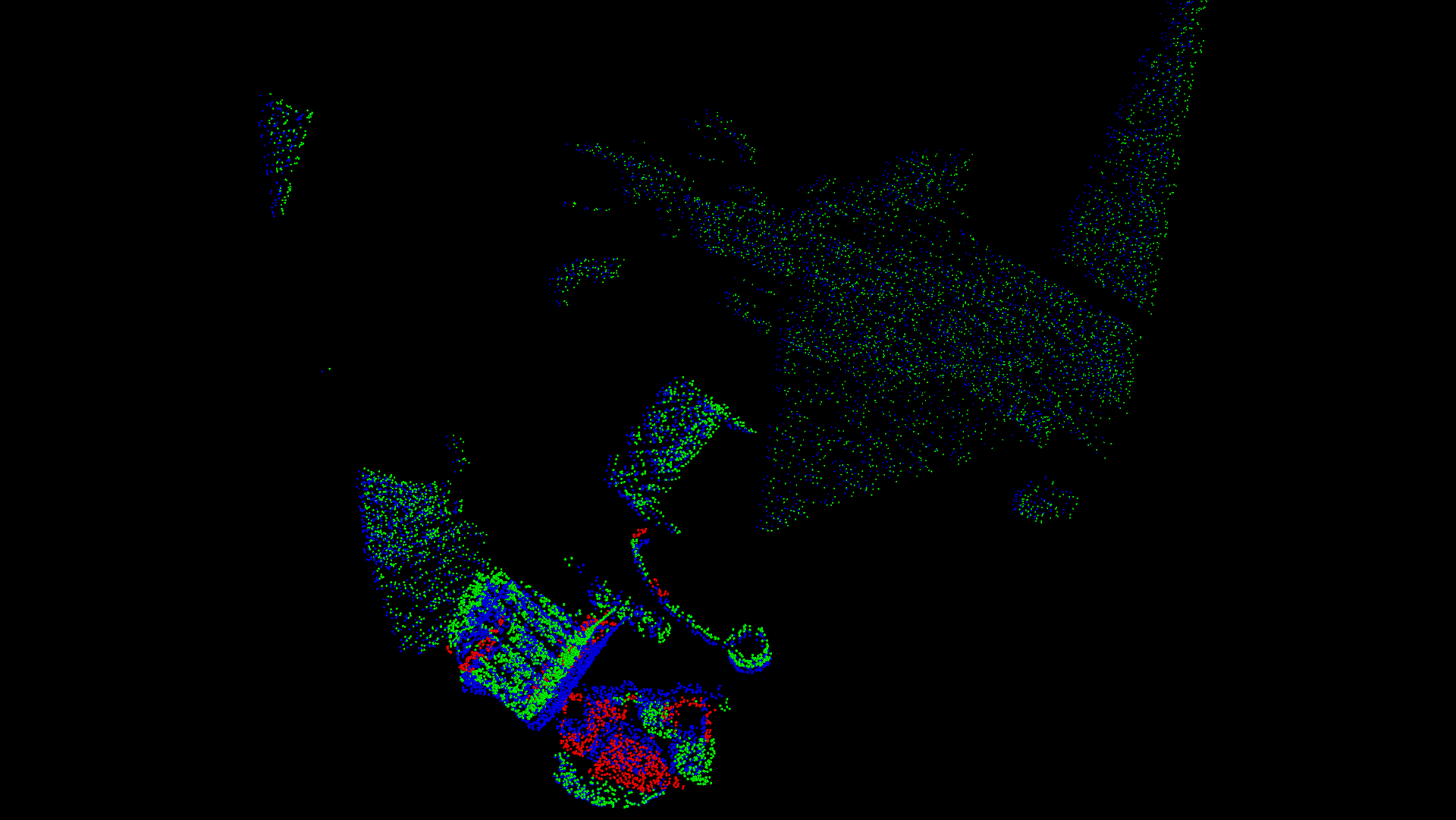} &
        \includegraphics[width=\sz\textwidth]{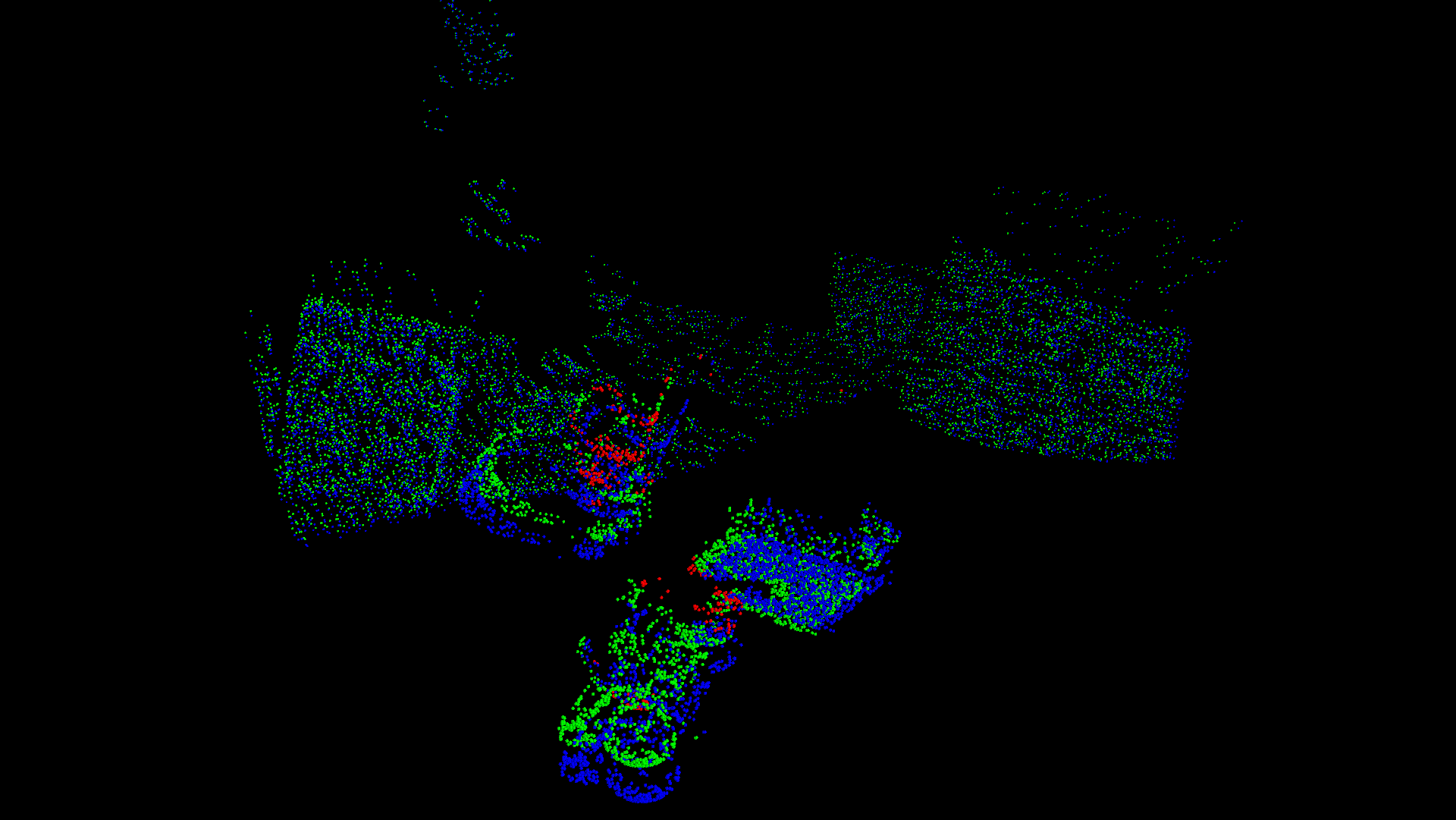} &
        \includegraphics[width=\sz\textwidth]{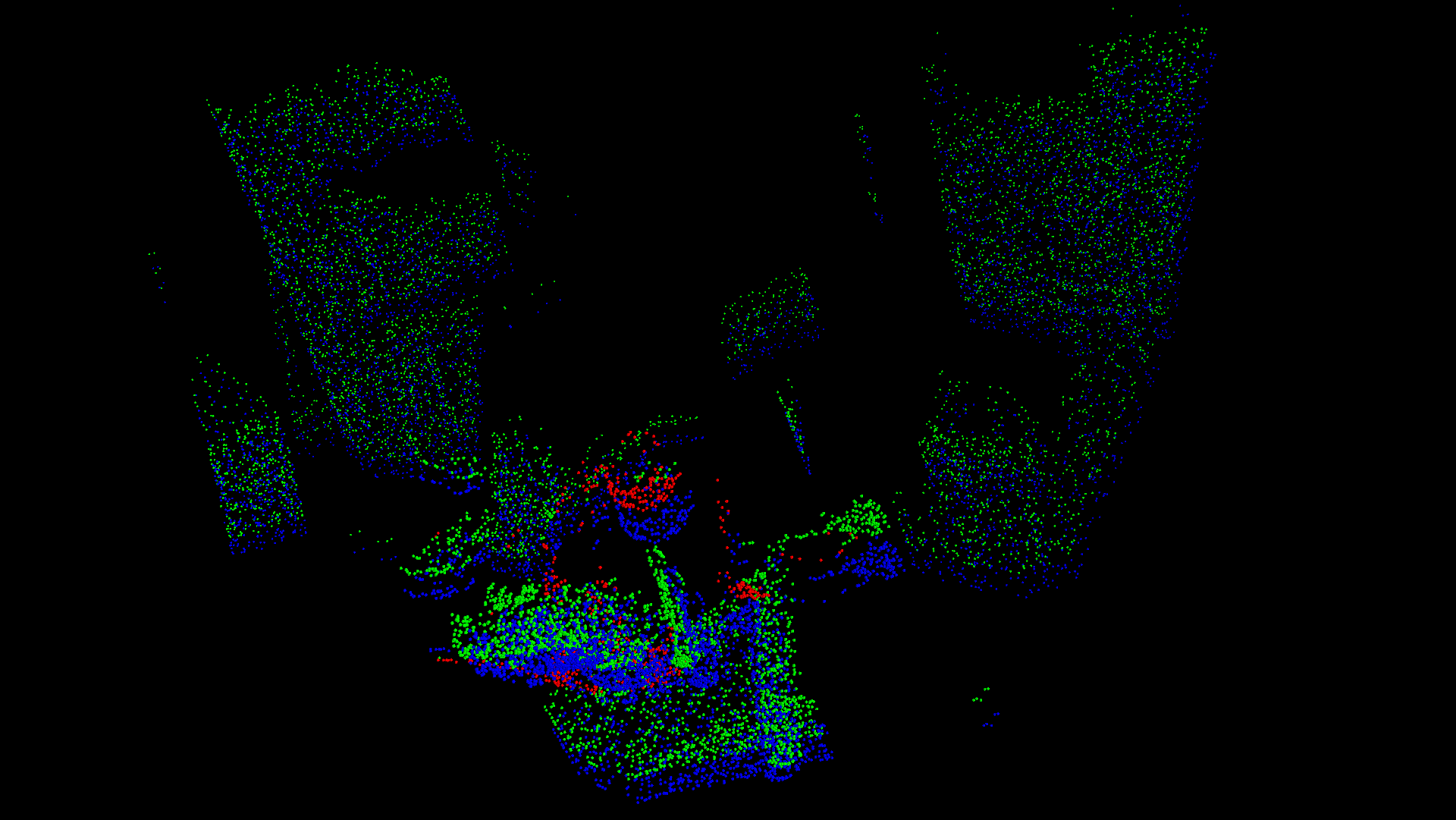} \\
        \multirow{1}{*}[50pt]{\rotatebox{90}{Self-Supervised}} &
        \includegraphics[width=\sz\textwidth]{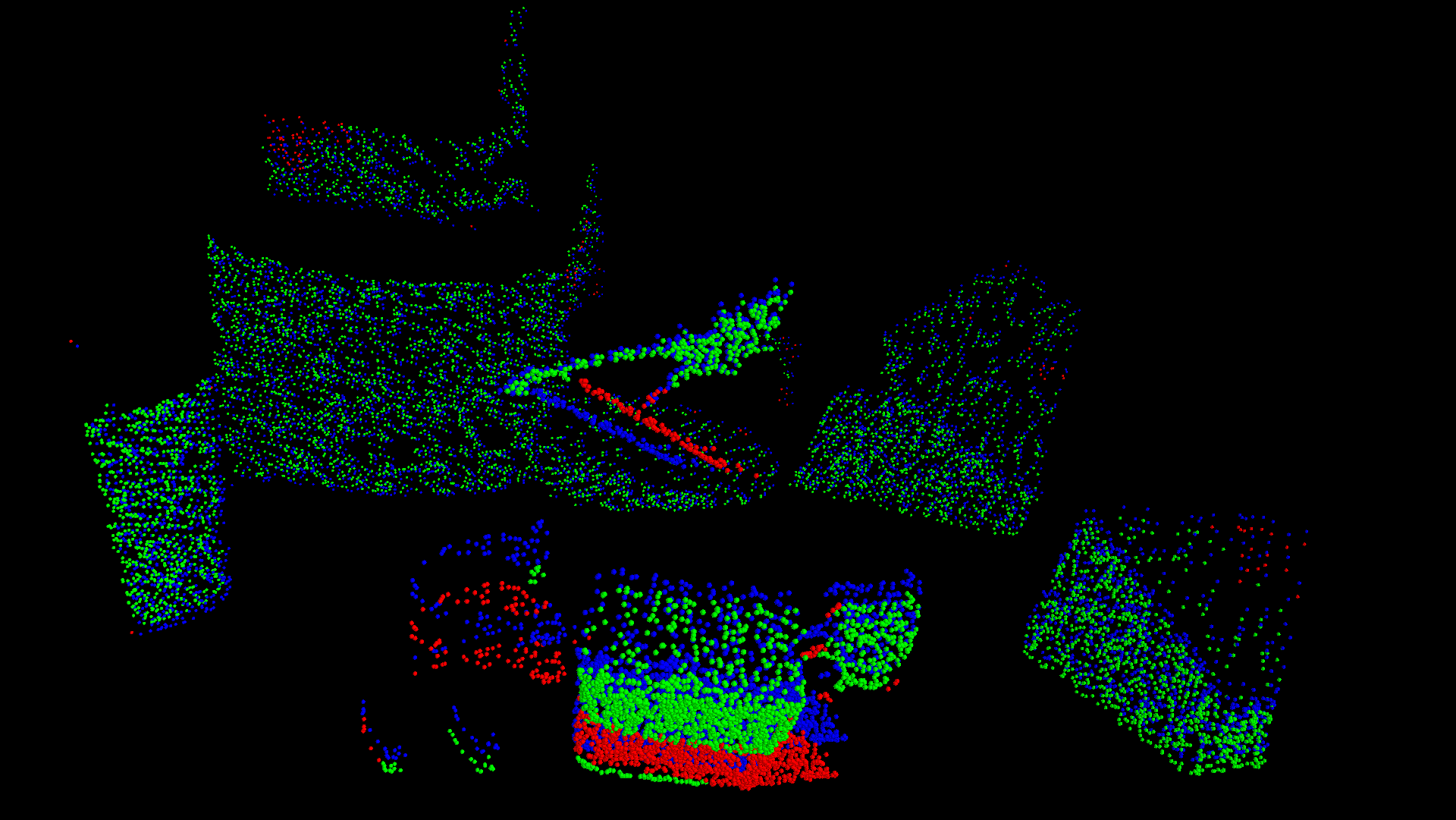} &
        \includegraphics[width=\sz\textwidth]{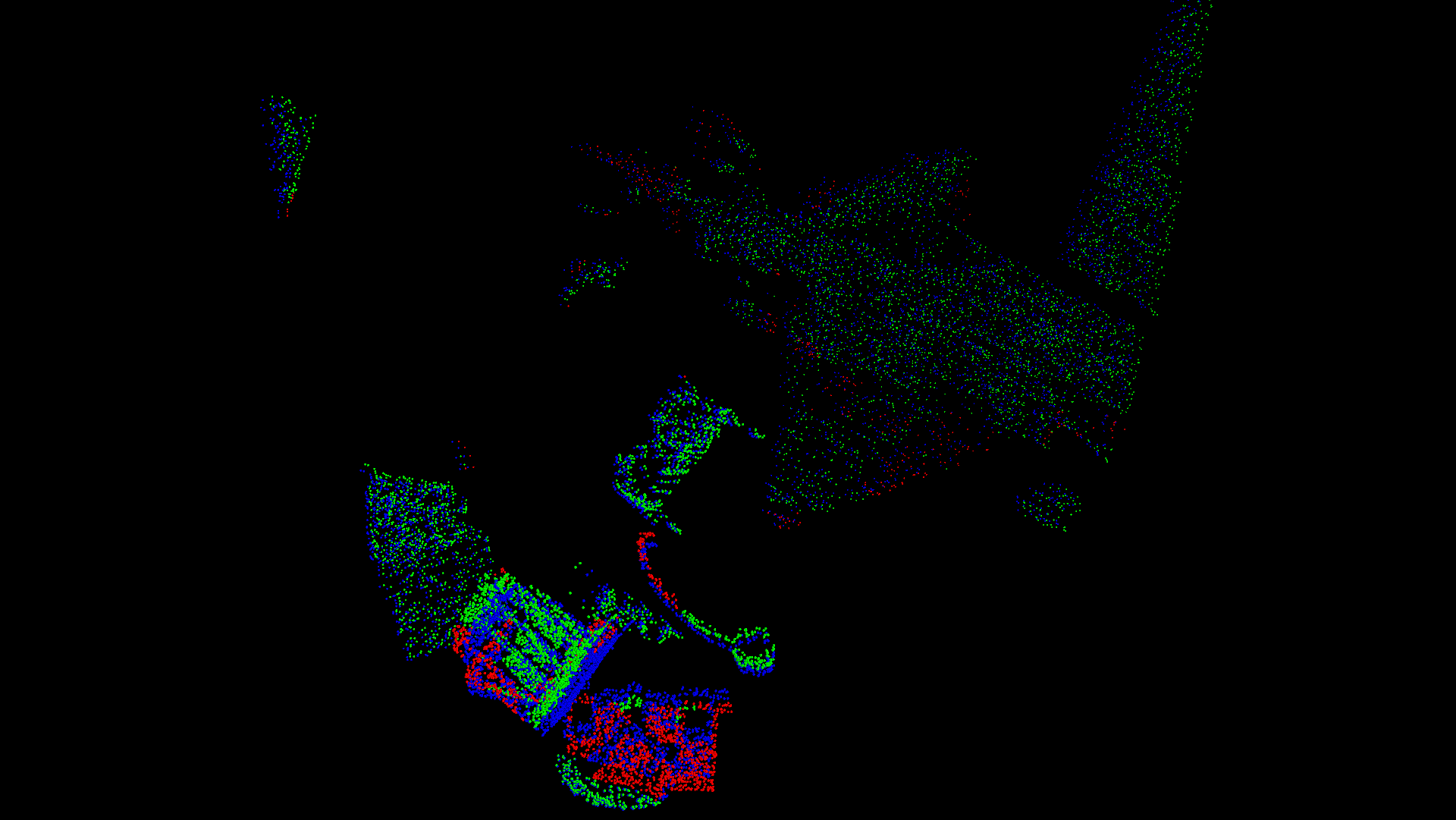} &
        \includegraphics[width=\sz\textwidth]{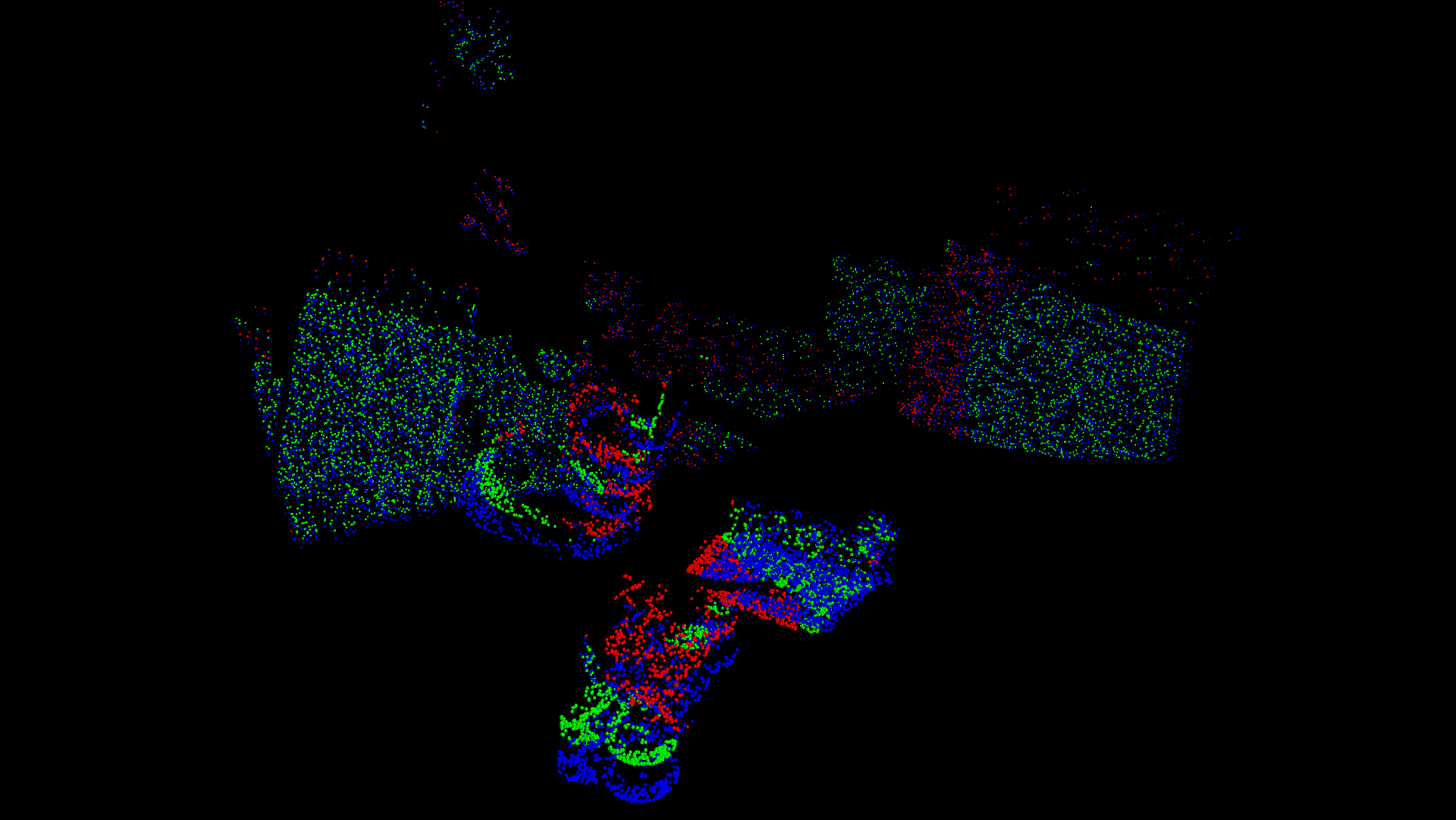} &
        \includegraphics[width=\sz\textwidth]{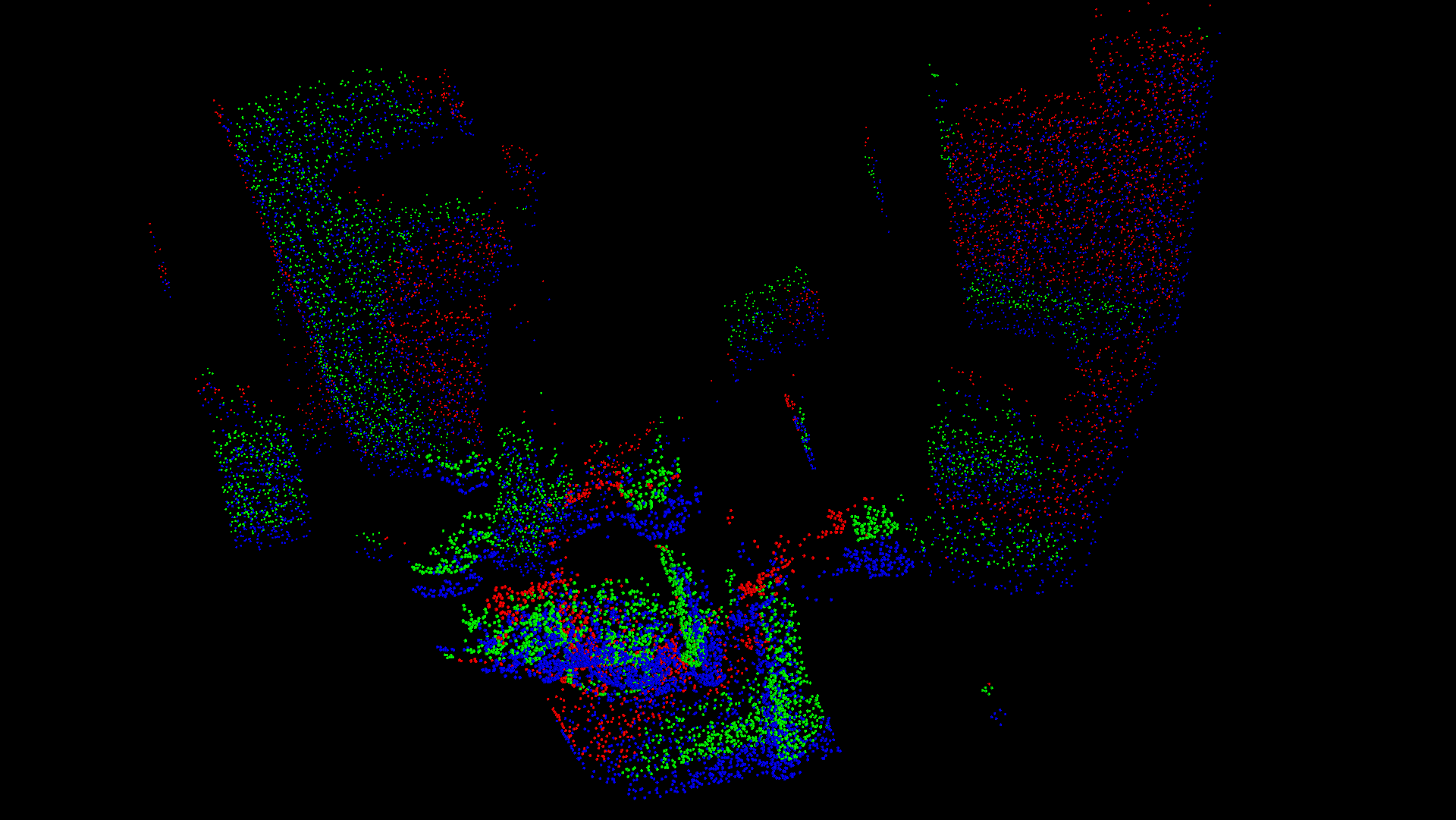} \\
        \multicolumn{5}{c}{FlyingThings3D} \\[+0.1cm]
        \multirow{1}{*}[35pt]{\rotatebox{90}{Hybrid}} &
        \includegraphics[width=\sz\textwidth]{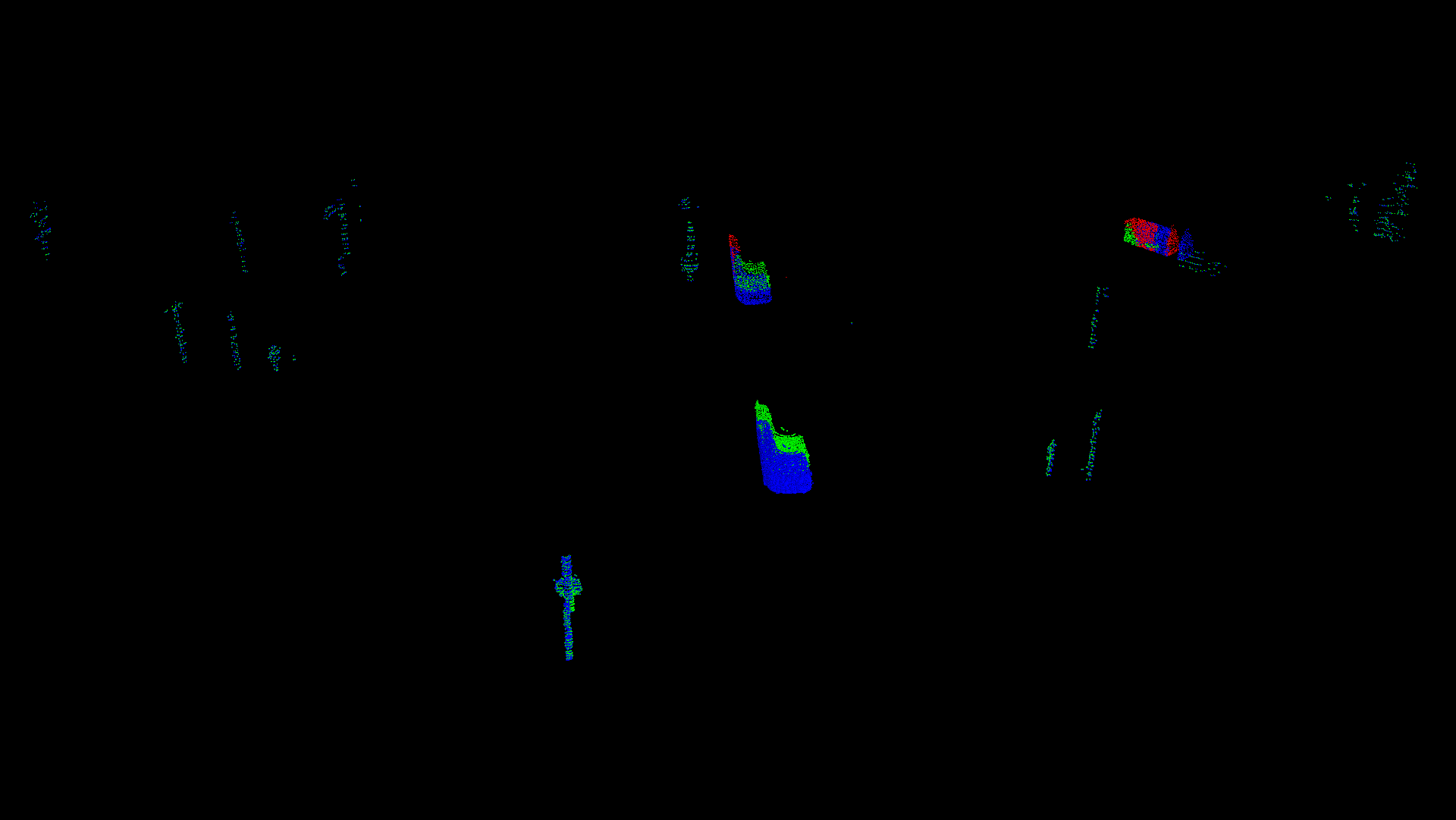} &
        \includegraphics[width=\sz\textwidth]{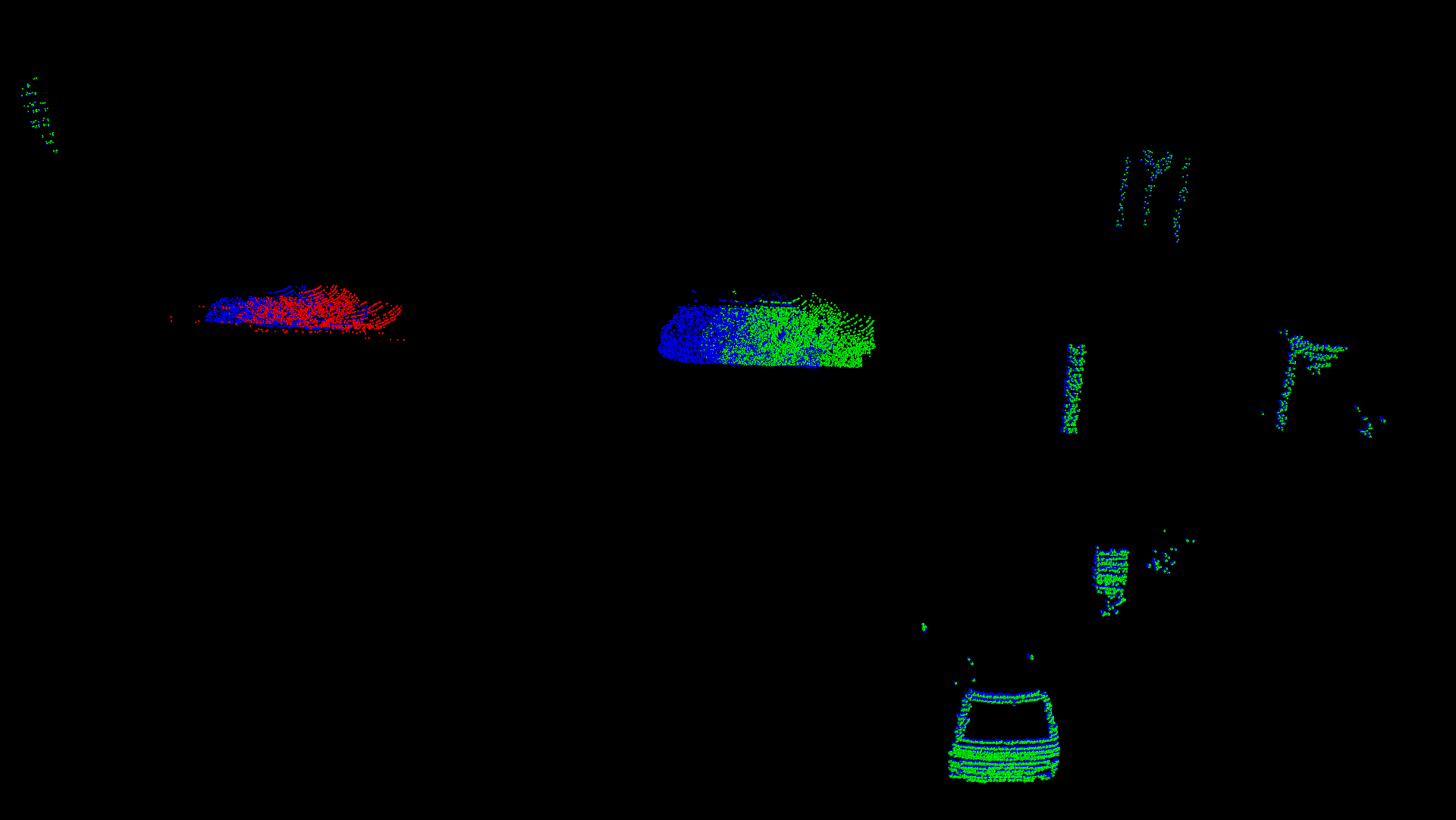} &
        \includegraphics[width=\sz\textwidth]{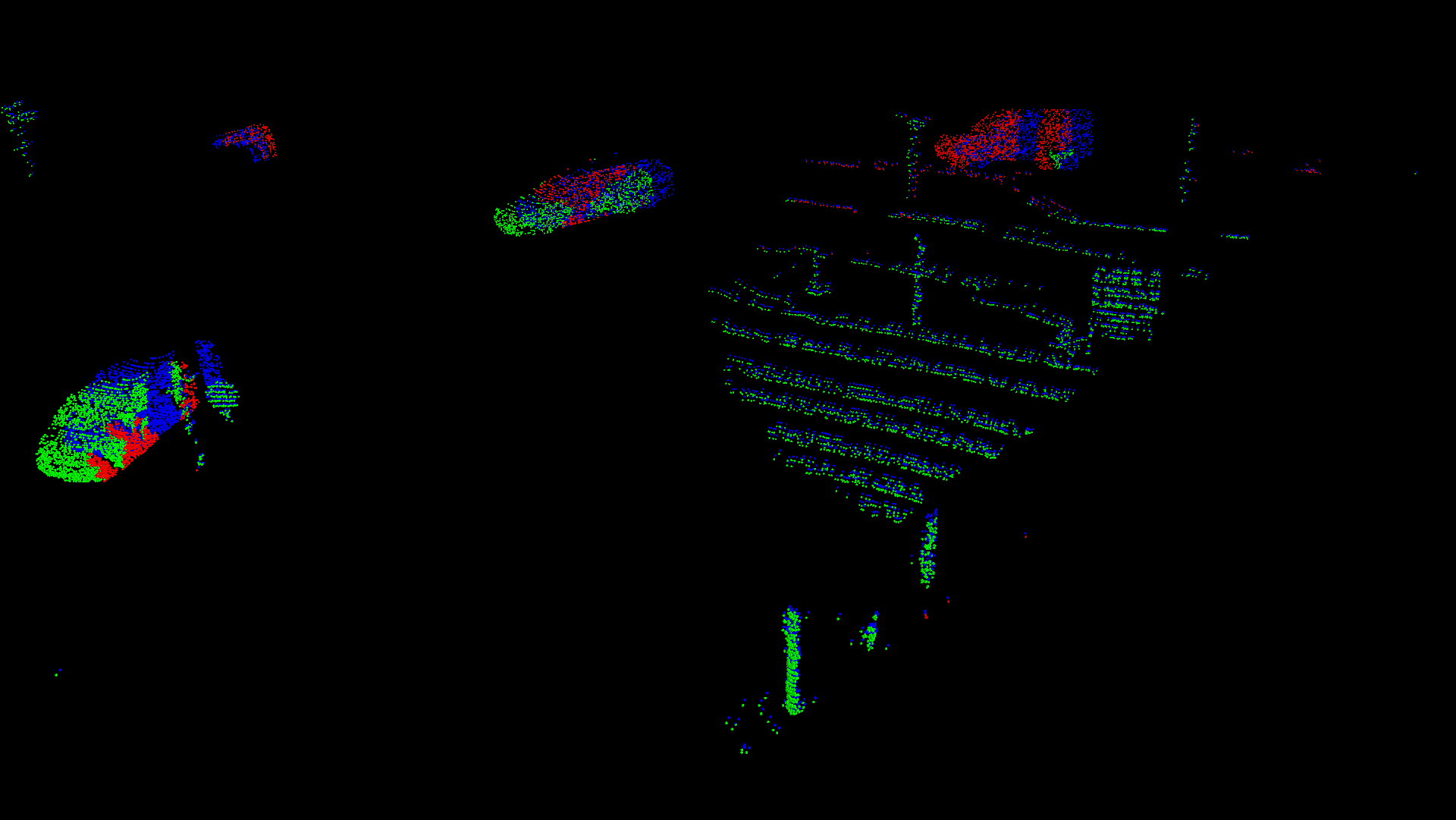} &
        \includegraphics[width=\sz\textwidth]{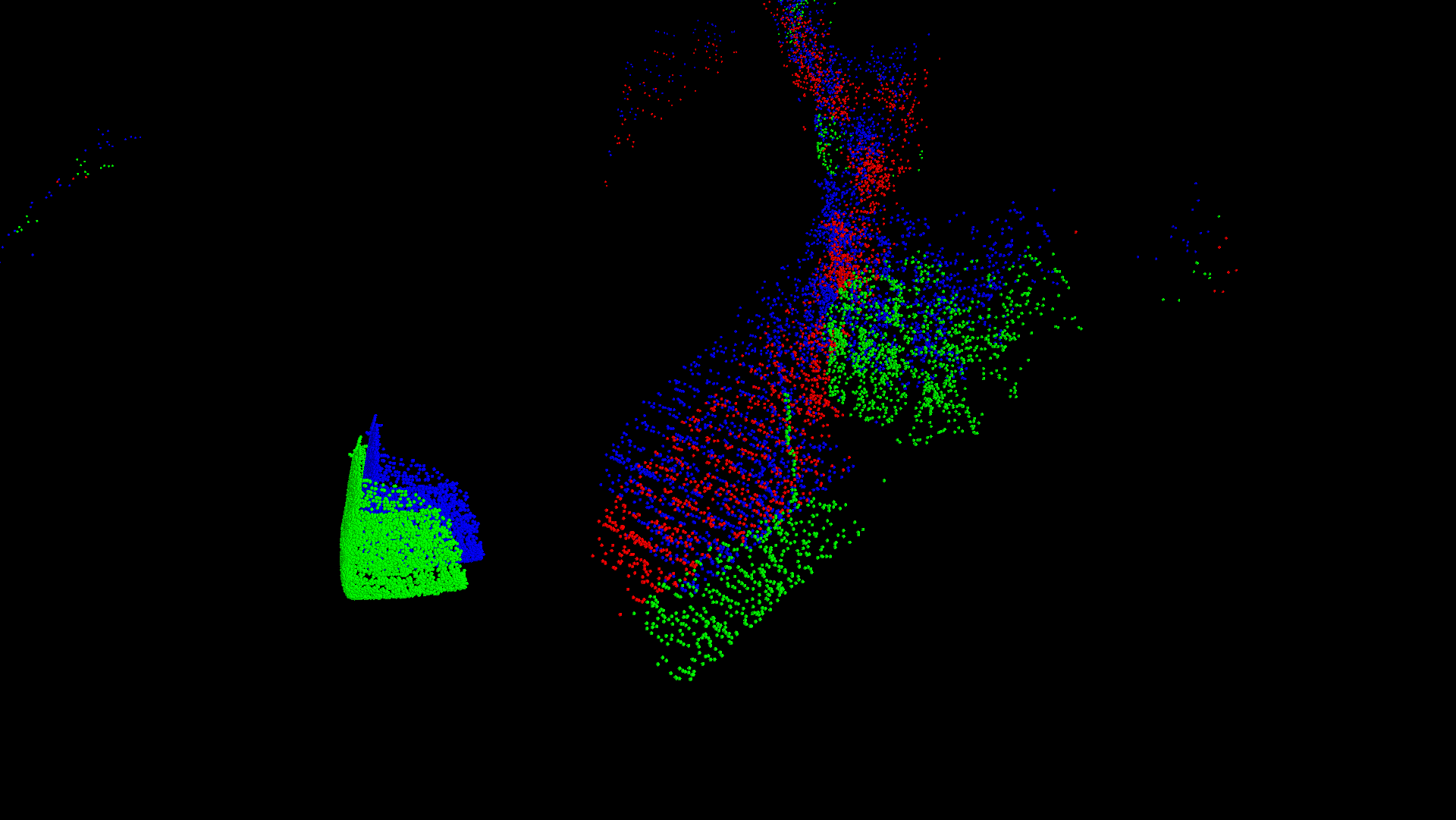} \\
        \multirow{1}{*}[50pt]{\rotatebox{90}{Self-Supervised}} &
        \includegraphics[width=\sz\textwidth]{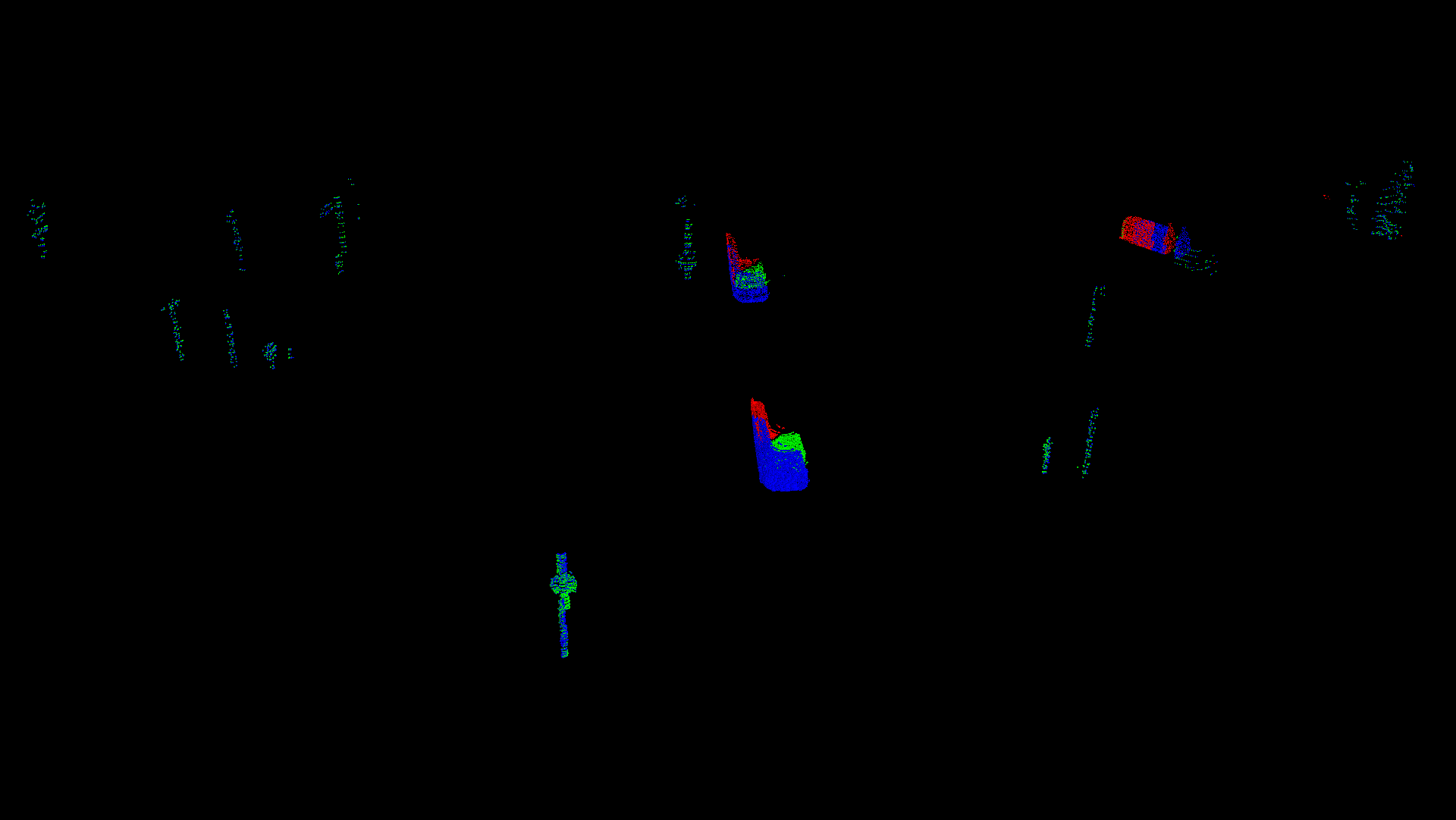} &
        \includegraphics[width=\sz\textwidth]{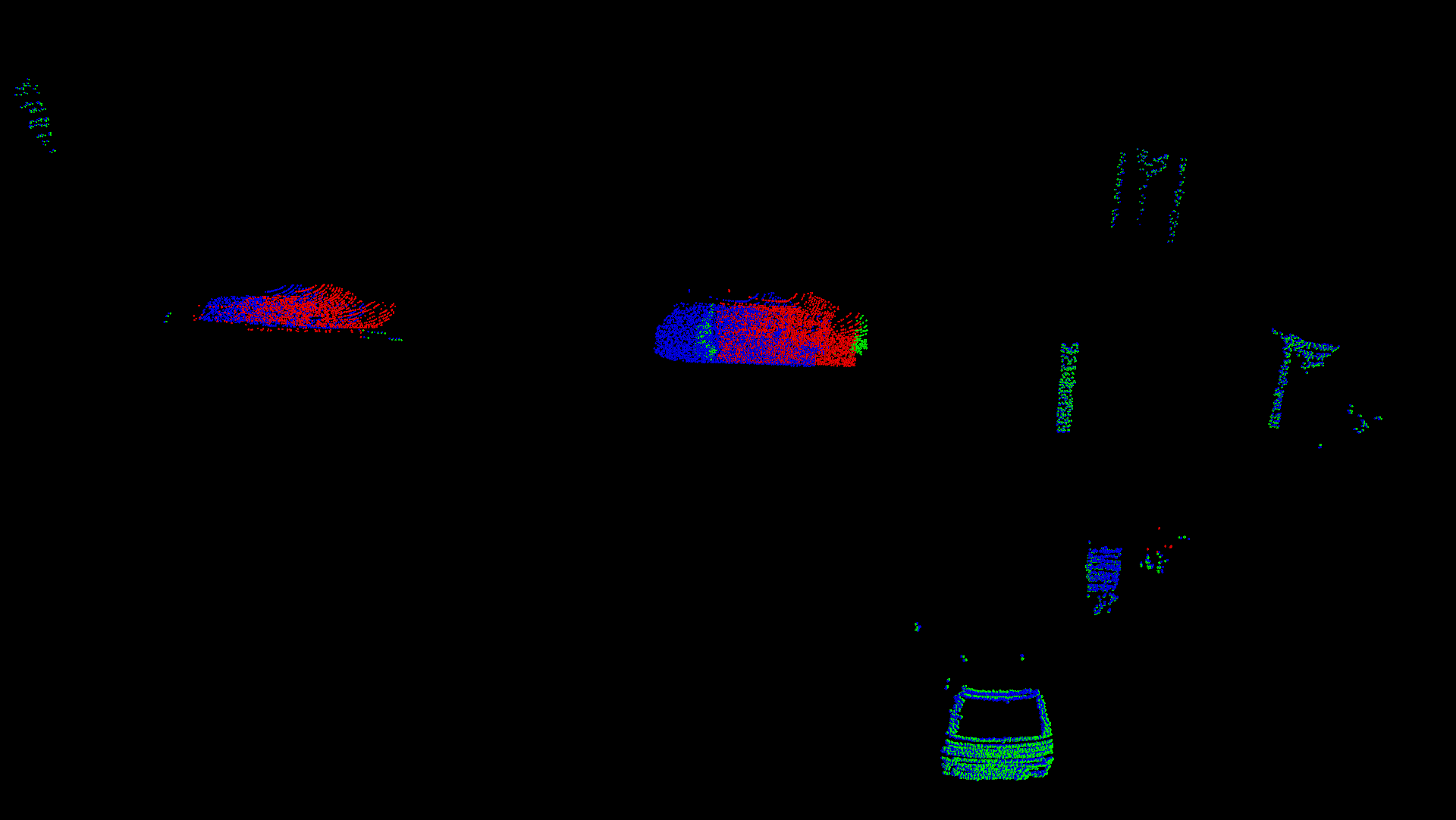} &
        \includegraphics[width=\sz\textwidth]{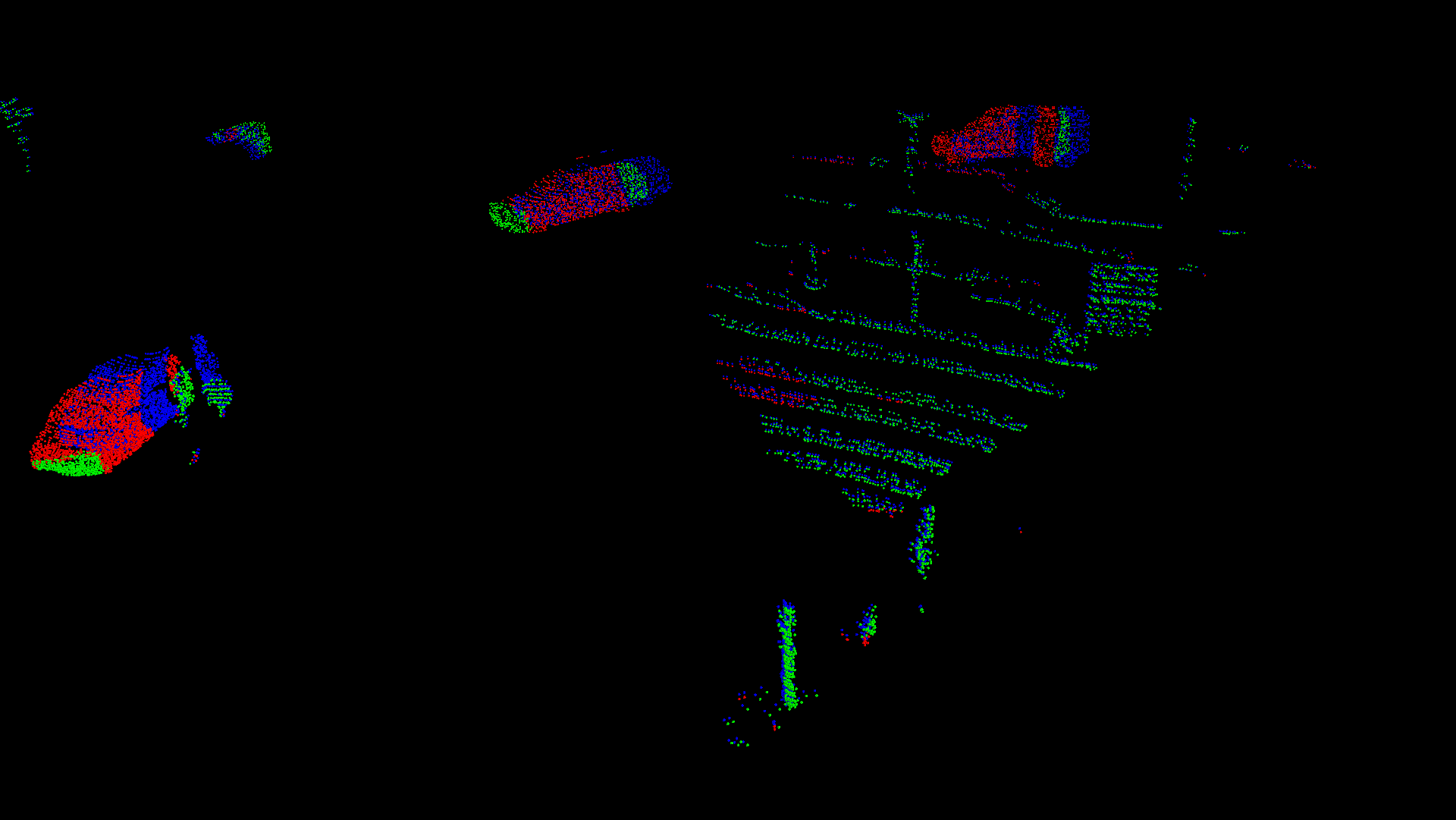} &
        \includegraphics[width=\sz\textwidth]{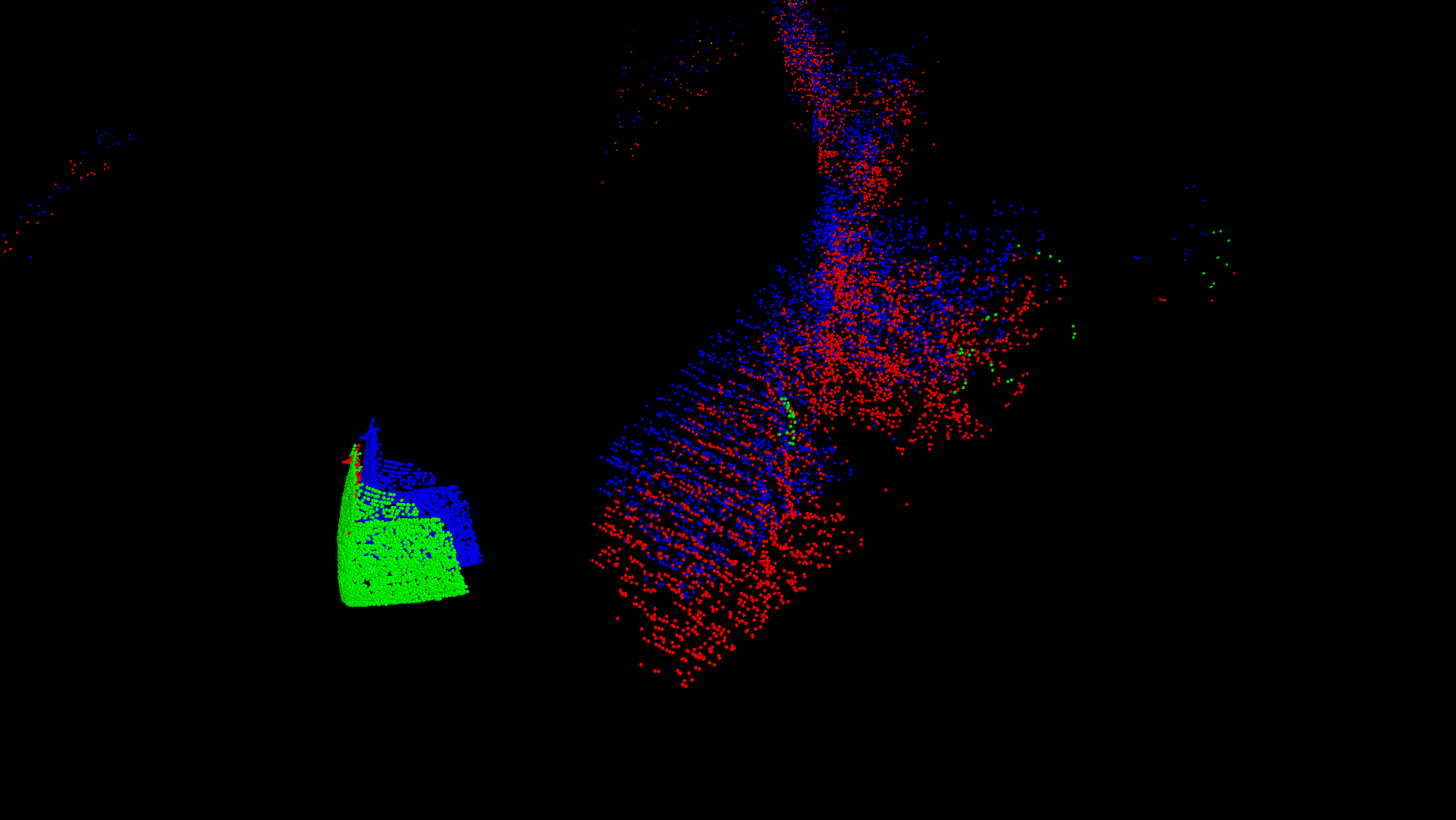} \\
        \multicolumn{5}{c}{KITTI} \\ [-0.2cm]
    \end{tabular}
\caption{\textbf{Qualitative results on FlyingThings3D~\cite{mayer2016large} and KITTI~\cite{Menze2015CVPR}.} For each dataset the top row depicts the results from hybrid training (supervised training with self-supervisory terms). The bottom row shows the results from fully self-supervised training. The original input point cloud $\pc{1}$ is displayed in blue. Correctly predicted points are shown in green as a warped point cloud with predicted total flows $\pc{1} + \hat{D}$. Wrongly predicted points are shown in red as points warped with the ground-truth total flows $\pc{1} + {D}$. Correctness is defined according to Acc3D(0.1).}
\label{fig:quality_fly_kitti_sup}
\end{figure*}

In this section, we show additional qualitative results evaluated on FlyingThings3D~\cite{mayer2016large}. To test for generalization we take the same hybrid and self-supervised models and evaluate them on a real world dataset KITTI~\cite{Menze2015CVPR}. Additional qualitative results evaluated on both datasets are shown in Figure~\ref{fig:quality_fly_kitti_sup}.

Overall, hybrid training provided better results compared to self-supervised training according to larger areas of green regions. Self-supervised model sometimes struggled with large rotations and incomplete shapes on FlyingThings3D, while still providing comparable overall results. Both training modes were able to generalize to real world data from KITTI. Hybrid model generalized better especially on unseen surfaces and incomplete shapes. Self-supervised fine-tuning on a target dataset may be used to improve the results of a self-supervised model which was trained on FlyingThings3D.

% ----------------------------------------------------------------------------
\clearpage
%%%%%%%%% BIBLIOGRAPHY
{\small
\bibliographystyle{ieee}
\bibliography{egbib}
}
\end{document}